\documentclass[review,3p,times,authoryear]{article}

\usepackage{amssymb}
\usepackage{amsmath}

\usepackage{booktabs}
\usepackage{float}
\usepackage{subcaption}
\usepackage{placeins}
\usepackage[a4paper,margin=1in]{geometry}
\usepackage{authblk}
\usepackage{natbib}
\usepackage{float}
\usepackage[hidelinks]{hyperref} 
\usepackage{xurl}               
\usepackage{graphicx}

\title{MB-DSMIL-CL-PL: Scalable Weakly Supervised Ovarian Cancer Subtype Classification and Localisation Using Contrastive and Prototype Learning with Frozen Patch Features}

\author[1]{Marcus Jenkins\thanks{Corresponding author. Email: \texttt{marcus.jenkins@uea.ac.uk}}}
\author[2]{Jasenka Mazibrada}
\author[2]{Bogdan Leahu}
\author[1]{Michal Mackiewicz}

\affil[1]{School of Computing Sciences, University of East Anglia, Norwich, Norfolk, NR4 7TJ, UK}
\affil[2]{Norfolk and Norwich University Hospital, Norwich, Norfolk, NR4 7UY, UK}

\date{}

\begin{document}
\maketitle

\begin{abstract}
The study of histopathological subtypes is valuable for the personalisation of effective treatment strategies for ovarian cancer. However, increasing diagnostic workloads present a challenge for UK pathology departments, leading to the rise in AI approaches. While traditional approaches in this field have relied on pre-computed, frozen image features, recent advances have shifted towards end-to-end feature extraction, providing an improvement in accuracy but at the expense of significantly reduced scalability during training and time-consuming experimentation. In this paper, we propose a new approach for subtype classification and localisation in ovarian cancer histopathology images using contrastive and prototype learning with pre-computed, frozen features via feature-space augmentations. Compared to DSMIL, our method achieves an improvement of 70.4\% and 15.3\% in F1 score for instance- and slide-level classification, respectively, along with AUC gains of 16.9\% for instance localisation and 2.3\% for slide classification, while maintaining the use of frozen patch features.
\end{abstract}

\section{Introduction}
\label{introduction}
Ovarian cancer accounts for approximately one-eighth of cancer diagnoses among women worldwide, with an estimated 300,000 new cases diagnosed annually. Of these cases, approximately two thirds prove fatal \citep{cancer_statistics}. This high mortality burden is largely attributable to diagnosis at advanced stages (III-IV), when prognosis is significantly poorer \citep{symptoms}. Early-stage detection remains challenging since presenting symptoms are nonspecific and frequently misattributed to benign gynaecological or gastrointestinal conditions. Furthermore, conventional screening strategies and blood-based biomarkers that have proven valuable for other cancers show limited effectiveness for detecting early-stage ovarian cancer \citep{symptoms}.

However, if the disease is captured early, histopathological examination supports detailed characterisation of tumour biology—including classification into histological subtypes—that is highly beneficial in guiding effective treatment strategies with favourable prognostic outcomes \citep{subtypes_and_treatment}. In addition, histological analysis facilitates the study of tumour progression, including the transitions from benign to borderline lesions and from borderline to malignant carcinoma. There are several recognised histological subtypes of ovarian carcinoma (OC) \citep{subtypes_and_treatment}. High-grade serous carcinoma (HGSC) is the most prevalent malignant subtype, accounting for approximately 70\% of cases, while less common subtypes include low-grade serous carcinoma (LGSC), endometrioid adenocarcinoma (EA), mucinous adenocarcinoma (MA), and clear cell carcinoma (CCC). Borderline tumours also play a critical clinical role, with the two key subtypes being serous borderline (SB) and mucinous borderline (MB) \citep{borderline_tumours}.

Growing diagnostic workloads exacerbate the challenge, with only 3\% of UK pathology departments reporting adequate staffing to meet clinical demand \citep{pathology_workforce}. Deep learning methodologies have therefore gained attention for assisting the review of histopathology slides. Existing approaches typically address classification, segmentation, or object detection tasks. Among these, semantic segmentation provides superior interpretability—a requirement in high-stakes clinical contexts—but generating pixel-level annotations is costly and labour-intensive. Additionally, the large spatial resolution of whole-slide images (WSIs) presents significant computational barriers for conventional image segmentation pipelines.

\subsection{Multiple Instance Learning for Whole Slide Images}
In MIL, a WSI is partitioned into smaller patches (instances), which are embedded independently and aggregated to produce a slide-level feature representation for classification. Early approaches used simple pooling operations for aggregation \citep{MILReview}, while later attention-based pooling enabled models to assign dynamic weightings to instances \citep{ABMIL, DSMIL, CLAM}. To maintain low computational demands during training with thousands of patches per slide, MIL frameworks have typically relied on frozen, precomputed patch features \citep{ABMIL, DSMIL, CLAM}. Although this offers rapid experimentation, performance is ultimately limited by the discriminative power of the frozen feature space. In turn, recent approaches have shifted towards training feature extractors end-to-end, improving task-specific instance discrimination but at the expense of a significant increase in computational demand and memory usage during training \citep{WENO, INS}.

\subsection{Cancer Localisation in MIL}
Localisation in MIL is traditionally performed using the heatmaps from attention-based aggregations directly. However, recent approaches have demonstrated superior performance using instance classifiers trained via pseudo-labelling \citep{CLAM, DSMIL, WENO, INS, DGMIL}. The precise method used for pseudo-labelling varies significantly between each approach; for example, attention maps in CLAM and WENO \citep{CLAM, WENO}, classifier confidence in DSMIL \citep{DSMIL}, negative-distribution modelling in DGMIL \citep{DGMIL}, and prototype similarity in INS \citep{INS}. Attention-based approaches can over-rely on a small subset of instances or highlight negative or contextual evidence, while confidence- or distribution-based methods rely on unstable early predictions that can lead to confirmation bias and error propagation. However, INS stands out for its use of momentum-updated pseudo-labels and class prototypes, which provide superior stability in pseudo-labelling compared to other approaches. Consequently, INS achieves higher performance on benchmark datasets \citep{INS}. We note that the majority of these approaches report their performance on binary classification datasets (cancerous versus non-cancerous), and so their performance in subtype classification/localisation is widely untested.

\subsection{Our Contributions}
In this work, we introduce a new weakly supervised combined slide and instance-classification approach that substantially improves class localisation using recent innovations while maintaining scalability and low computational overhead with precomputed patch feature extraction. Using only slide-level annotations for training, we evaluated two key established MIL frameworks alongside our proposed method for classifying and localising normal, borderline, and malignant ovarian tissue subtypes in WSIs.

The remainder of this paper is organised as follows. Section \ref{sec:method} details the baseline models, data preprocessing, and our proposed approach. Section \ref{sec:experiments} presents a comprehensive evaluation and the findings of our experiments, and Section \ref{sec:conclusion} concludes with a summary of our contributions and potential future work.

\section{Method} \label{sec:method}

\subsection{Baseline Approaches} \label{sec:baseline_selection}
CLAM \citep{CLAM} and DSMIL \citep{DSMIL} were selected as baseline MIL approaches, since they both utilise frozen, pre-computed patch embeddings for slide classification and cancer localisation. Both aggregate instances via attention but differ in pseudo-labelling. CLAM assigns labels from attention scores, while DSMIL selects the most confident instances; these represent two well-established and distinct approaches. CLAM and DSMIL are discussed in more depth in Sections \ref{sec:clam} and \ref{sec:dsmil}, respectively.

\subsection{Patch Extraction and Preprocessing} \label{sec:embedding}
Although our baseline methods use different combinations of patch sizes, magnifications, and patch encoders---DSMIL uses a SimCLR-pretrained ResNet-18 with 224$\times$224 patches at 5$\times$/20$\times$, and CLAM uses UNI on 256$\times$256 patches with a 25\% overlap---we use the same patch extraction and embedding pipeline to ensure that architectural comparisons are not confounded by differences in preprocessing. The following sections detail our patch extraction and preprocessing approach.

\subsubsection{Background and Tissue Segmentation} \label{sec:tissue_segmentation_method}
Each slide contains background regions that lack meaningful information conducive to the learning process, and thus including these would result in unnecessary computational overhead. To remove background tissue, we use preliminary saturation thresholding as described in \cite{CLAM} to identify tissue regions from which patches should be extracted. We first convert the slide to the HSV colour space, then a median blur is applied with a kernel size of 7 to reduce noise. The tissue regions are then identified using a fixed saturation threshold of 20. However, we did not perform the additional refinement step in \cite{CLAM} where contours are used to identify primary tissue regions and holes, since this yielded mixed results. In some cases, the subtraction of inner contours resulted in the removal of regions of adipose tissue, while in others the inclusion of contours led to the loss of valid tissue holes.

\subsubsection{Patch Extraction and Embedding}
We divide each WSI into non-overlapping patches of size 224$\times$224 pixels at 10$\times$ magnification from areas of tissue identified via the initial tissue segmentation as described in Section \ref{sec:tissue_segmentation_method}. This patch size was chosen since it aligns with the native input resolution of UNI \citep{UNI}, which we selected as our pretrained patch encoder for all experiments. Similarly, we use a magnification of 10\(\times\) based on the findings of \citet{breen_multi_resolution} and \citet{wolflein2023benchmarking}, where they report the optimal standalone magnification to be 10\(\times\). However, in \citet{breen_multi_resolution}, they demonstrate that using both 10\(\times\) and 20\(\times\) magnifications offers significant improvement, but we do not use multiple resolutions in this work due to the significant increase in computational complexity.

We use the UNI vision transformer as our patch encoder due to its strong performance in recent approaches: the graph-based approach of \citet{breen_multi_resolution} and in the latest iterations of CLAM \citep{CLAM_github}. Furthermore, \citet{breen_multi_resolution} demonstrate that vision transformers (including UNI) consistently outperform CNN-based patch encoders in all experiments. UNI is trained on over 100 million histopathological images using DINOv2 \citep{DINO}, which is a self-supervised approach that produces semantically meaningful and augmentation-invariant embeddings through self-distillation. These properties make UNI particularly suitable for weakly supervised learning tasks where quality of feature space representation is crucial.

\subsection{DSMIL} \label{sec:dsmil}
DSMIL \citep{DSMIL} consists of two streams, an instance classifier and a bag classifier, and uses critical instance pseudo-labelling (the most confident instance per class from the instance classifier) to guide both the supervision of the instance classifier and bag-level aggregation. Specifically, attention scoring is computed between all patches and the critical instance during aggregation, thus assigning a higher attention score to instances that are more similar to the most probable positive instance. However, a current limitation of DSMIL is that it is designed for binary or multi-label tasks, rather than multi-class classification. To adapt DSMIL for multi-class subtype classification, we first replace the bag-level loss with cross-entropy and adapt the critical instance selection strategy.

In DSMIL, the critical instance \( h_m^c \) for each class \( c \in \{0, \dots, C{-}1\} \) is selected via:
\begin{equation}
h_m^c(\mathcal{B}_i) = \arg\max \left\{ W_c h_{i0}, \dots, W_c h_{i(N{-}1)} \right\}
\end{equation}
where \( W \in \mathbb{R}^{C \times d} \) is the instance classifier and \( h_{ij} \) are the instance features.

We adapt this for multi-class classification by separating subtype classes \( P_c \in \{0, \dots, C{-}2\} \) and the normal class \( N_c = C{-}1 \), selecting:
\begin{align}
h_m^{P_c}(\mathcal{B}_i) &= \arg\max \left\{ W_{P_c} h_{i0}, \dots, W_{P_c} h_{i(N{-}1)} \right\} \\
h_m^{N_c}(\mathcal{B}_i) &= \arg\max \left\{ W_{N_c} h_{i0}, \dots, W_{N_c} h_{i(N{-}1)} \right\}
\end{align}

Each \( h_m^{P_c} \) is supervised using the slide label, while \( h_m^{N_c} \) is supervised using the normal class label. This penalises high-confidence predictions for classes not present in the slide. For normal slides, all critical instances \( h_m^c \; \forall c \in \{0, \dots, C{-}1\} \) are assigned the normal class label.


\subsection{CLAM} \label{sec:clam}
In CLAM~\citep{CLAM}, each class-specific attention branch computes attention scores \( a_j^c \) over instance embeddings \( h_{ij} \) for class \( c \), giving the attention vector \( A^c \). These scores are used in instance aggregation to provide a class-specific bag representation:
\begin{equation}
M^c = \sum_j a_j^c \cdot h_{ij}
\end{equation}
which is passed through a class-specific linear classifier to produce the bag-level logit for class \( c \). For the slide's ground truth class, the top-$k$ and bottom-$k$ instances based on \( A^c \) are assigned positive and negative pseudo-labels. For all other classes, only the top-$k$ instances are sampled from their respective attention map (\(A_c\)) and are assigned negative labels. As such, the logits for each instance are provided in a one-versus-rest format, where there are positive and negative logits for each class \( c \), denoted as \( \ell_{c,1} \) and \( \ell_{c,0} \), respectively. To optimise the instance classifier, CLAM proposes using either cross-entropy or smooth SVM loss~\citep{smoothsvm}, which encourages margin-based separation rather than optimising logits to hard targets. In preliminary experimentation, we found negligible difference in performance between these loss functions, and so we use cross-entropy loss for consistency with DSMIL. 

A limitation of CLAM is that it does not support training with negative (normal tissue) slides. Therefore, we add a normal class branch to the bag classifier, but not to the instance classifier. For normal slides, we apply the out-of-class sampling strategy to all classes. During inference, the slide is predicted as normal if all \( \ell_{c,0} > \ell_{c,1} \); otherwise, the predicted class is the one with the highest positive logit:

\begin{equation}
\hat{y} =
\begin{cases}
C_{\text{normal}}, &
\begin{aligned}
\text{if } & \arg\max_{j \in \{0,1\}} \ell_{c,j} = 0 \\
           & \forall \; c \in \{0, \dots, C{-}1\}
\end{aligned} \\
\arg\max_{c} \ell_{c,1}, & \text{otherwise}
\end{cases}
\label{eq:prediction_rule}
\end{equation}

\subsection{Proposed Method (MB-DSMIL-CL-PL)}
We further DSMIL's performance for slide and instance classification by integrating aspects of CLAM and INS \citep{INS}, while maintaining support for precomputed instance embeddings in a novel manner. The following sections detail each proposed adaptation.

\subsubsection{Multi-Branch DSMIL (MB-DSMIL)}
The adaptation of DSMIL proposed in \cite{DSMIL} for multiple class settings employs a shared query projection function for instance aggregation for the bag feature vector. However, CLAM uses class-specific attention branches and classifiers, which enable higher expressive capacity to capture class-specific relevance of instances to the slide prediction. Therefore, we introduce Multi-Branch DSMIL (MB-DSMIL), which incorporates class-specific query projections for the attention mechanism of DSMIL.

Formally, let $\mathcal{B}_i = \{h_{ij}\}_{j=1}^{N_i}$ denote a bag of instance features, and let \(c \in C\) represent the possible classes. For each class $c$, we define a class-specific query projection function for \(h_{ij}\), $\phi_c: \mathbb{R}^d \to \mathbb{R}^q$. Then, the self-attention mechanism computes the softmax attention between the query embeddings of all instances and the query embedding of the critical instance for class \(c\) through $\phi_c$. The resultant bag feature vector after weighted aggregation is specific to class \(c\), and is passed through a class-specific classifier that produces the logit for class \(c\).

\subsubsection{Contrastive Learning (CL)}
INS applies a supervised contrastive loss \citep{supcon} to encourage clustering of instance features based on pseudo-labels. The loss is computed between weak and strong augmentations of each instance, as well as between instances with shared pseudo-labels. To do so, INS employs a query encoder \( f_q \) for weak views and a key encoder \( f_k \) for strong views, producing embeddings \( q_{ij} \) and \( k_{ij} \). The key encoder is updated via exponential moving average (EMA) of \( f_q \). A memory queue of up to 8,192 previous \( k_{ij} \) vectors with their pseudo-labels is also used to enable contrastive learning across a larger pool. During the first 80 epochs, an unsupervised MoCo-style contrastive loss \citep{moco} is used to align weak and strong views in the absence of reliable pseudo-labels.

However, in INS, \( f_q \) is a ResNet-18 model and is trained end-to-end with images patches, leading to significant memory demands and computational complexity. Instead, we adapt INS's contrastive learning to work with our precomputed instance features from UNI by applying the feature-space augmentations used in SimCL \citep{simcl}, and using a simple Multi-Layer Perceptron (MLP) layer for the query/key encoders. This allows for task-specific adaptability of the feature space of our general-purpose feature extractor (UNI), a benefit of end-to-end contrastive learning, without requiring costly end-to-end training in the image space. Specifically, SimCL augmentation adds normalised Gaussian noise constrained to the same hyperquartet as the original feature vector (i.e., maintaining the sign of each feature dimension) to prevent large semantic drift. Given an instance embedding \( h_{ij} \in \mathbb{R}^d \) and noise scaling factor \(\eta\), the noise-augmented feature \( h_{ij}' \) is computed as:
\begin{equation} \label{eq:simcl}
\begin{aligned}
\bar{\Delta} &\sim U(0, 1)^d \\
\Delta &= \eta \cdot \frac{\bar{\Delta} \odot \mathrm{sign}(h_{ij})}{\|\bar{\Delta}\|_2} \\
h_{ij}' &= h_{ij} + \Delta
\end{aligned}
\end{equation}

\begin{figure}[H]
    \centering
    \includegraphics[width=1\linewidth]{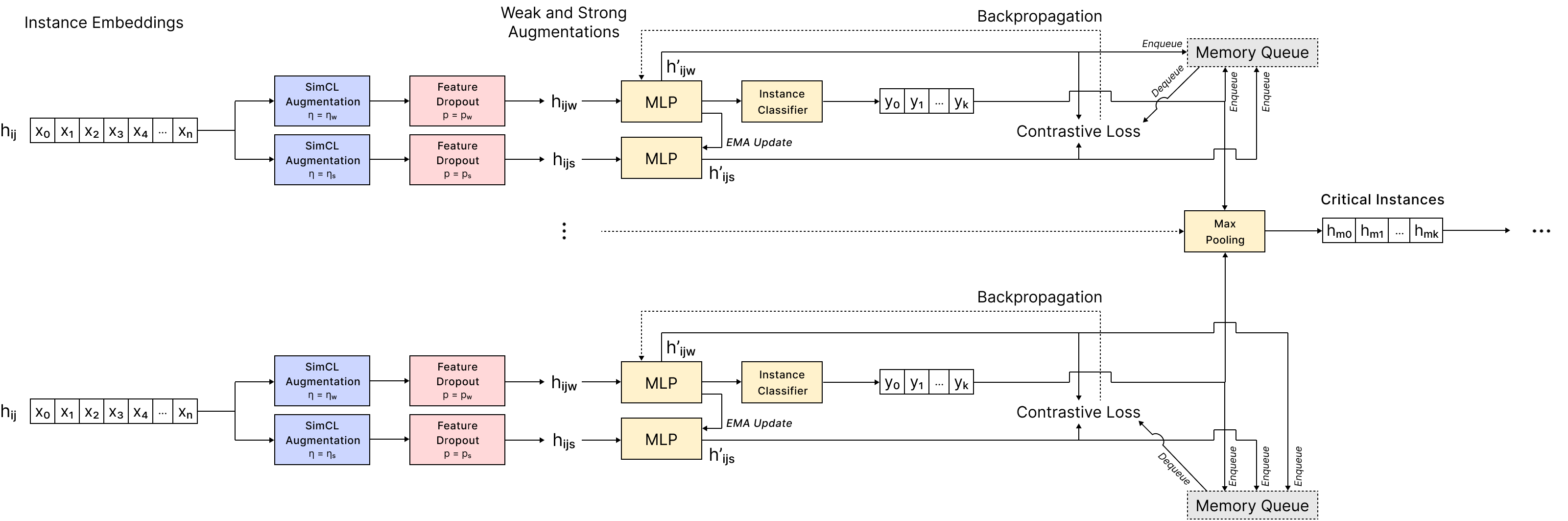}
    \caption{Architecture of the MB-DSMIL-CL contrastive learning. Instance classifier predictions
 are not used for the MoCo warmup epochs}
    \label{fig:mbdsmil-cl-architecture}
\end{figure}

We produce weak and strong instance augmentations \( h_{ijw} \) and \( h_{ijs} \) using scaling factors \( \eta_w \) and \( \eta_s \), with \( \eta_s > \eta_w \). We also evaluate feature dropout after SimCL augmentation, applying dropout probabilities \( p_w \) and \( p_s \) independently to \( h_{ijw} \) and \( h_{ijs} \). These are then projected via the MLP layers \( g_q, g_k: \mathbb{R}^d \rightarrow \mathbb{R}^e \) to give \(h_{ijw}'\) and \(h_{ijs}'\), where \( g_k \) is a momentum-updated version of \( g_q \). We use MoCo-style contrastive loss for a number of warm-up epochs, followed by supervised contrastive loss. The approach used to assign pseudo-labels to \( h_{ijw}' \) (and the corresponding \( h_{ijs}' \)) for contrastive loss is described in the following section. Figure \ref{fig:mbdsmil-cl-architecture} illustrates our proposed architecture. To our knowledge, this is the first approach to apply contrastive loss within a MIL framework using SimCL augmentation on precomputed features.

\subsubsection{Prototype Learning (PL)}
INS \citep{INS} introduces class prototype learning for instance classification by assigning soft pseudo-labels based on the similarity between the weakly augmented instance feature vector (equivalent to our \(h_{ijw}'\)) and class prototypes \( \mu_0 \) and \( \mu_1 \) for the negative and positive classes, respectively. The prototypes are initialised with Gaussian noise and updated via exponential moving average (EMA) using instance features conditioned on the predicted label \( \hat{y} \in \{0, 1\} \):

\begin{equation}
\mu_{\hat{y}} \leftarrow \text{norm}(m \cdot \mu_{\hat{y}} + (1 - m) \cdot h_{ijw}')
\end{equation}

where \( m = 0.9 \), and \( \text{norm} \) denotes \(\ell_2\) normalisation.

To improve stability, soft pseudo-labels \( s_{ij} \) for positive instances are updated using a momentum-based approach:

\begin{equation}
\begin{aligned}
z_{ij} &= \text{onehot}\!\left(\arg\max_{c \in \{0,1\}} \text{norm}(h_{ij})^\top \mu_c\right) \\
s_{ij} &\leftarrow \alpha \cdot s_{ij} + (1 - \alpha) \cdot z_{ij}
\end{aligned}
\end{equation}

where \( \alpha = 0.8 \). For negative-slide instances, \( s_{ij} \) is fixed as the negative label. For positive-slide instances, \( s_{ij} \) is initialised with class priors based on slide-level labels. The instance classifier is trained using KL divergence between predicted logits and \( s_{ij} \).

We extend this prototype-based pseudo-labelling to MB-DSMIL combined with contrastive learning by replacing the supervision of critical instances with supervision based on the pseudo-labels from prototype similarity. Unlike INS, which uses only positive and negative prototypes, we define a prototype \( \mu_c \) for each class \( c \in \{0, 1, \dots, C{-}1\} \), including the normal tissue. The \(\arg\max\) operations are restricted to the slide label and normal tissue class, and prototype similarity is computed for our \(h_{ijw}'\) produced via feature-space augmentation and projection through \(g_q\).

\begin{figure}[H]
    \centering
    \includegraphics[width=1\linewidth]{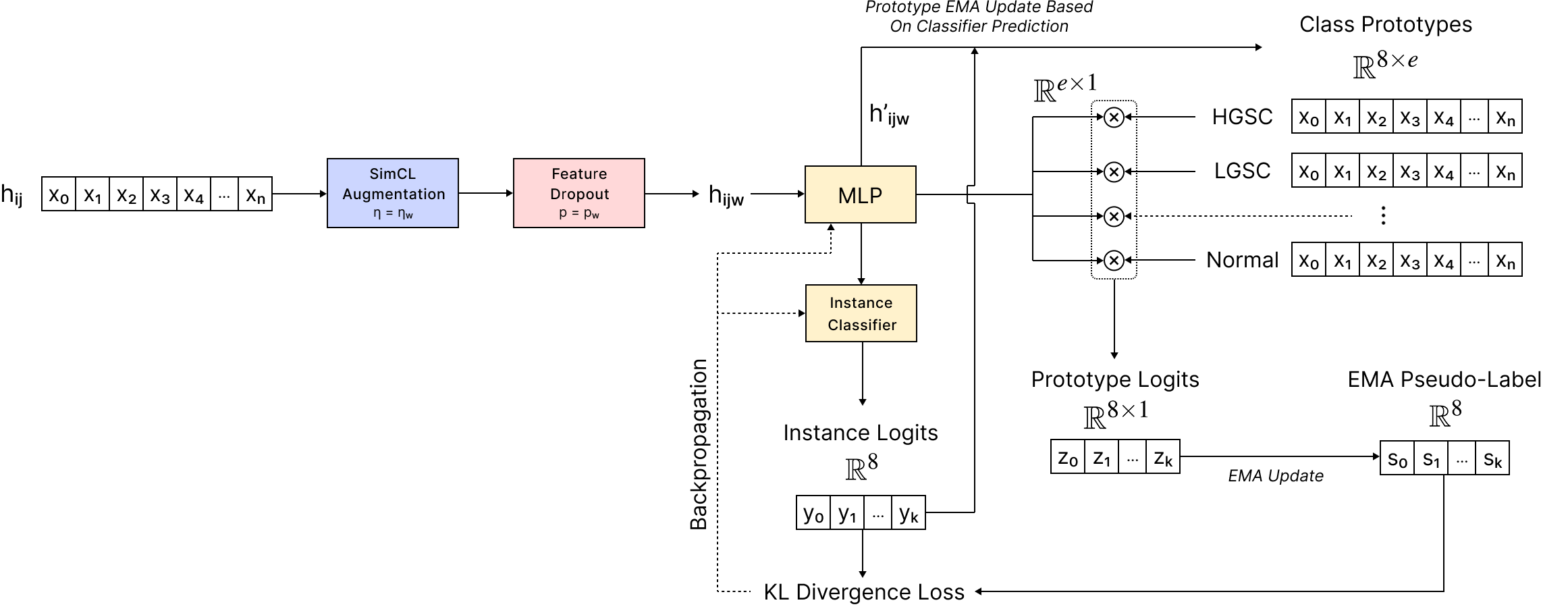}
    \caption{Class prototype learning with precomputed instance features in MB-DSMIL-CL-PL.}
    \label{fig:mbdsmil-cl-pl-architecture}
\end{figure}

\subsection{Experiments} \label{sec:experiments}
The following sections start by discussing our dataset, then our training and evaluation methodology. We finish by detailing our experiments and results for DSMIL, CLAM, and our proposed method, MB-DSMIL-CL-PL.

\subsubsection{Dataset}
Experimentation was performed using the DROV dataset \citep{DROV}. The DROV dataset comprises 174 histopathology slides of ovarian carcinoma scanned at 20× magnification with polygonal annotations for malignant and borderline subtypes, and normal tissue slides. Specifically, slides were scanned using a Scanscope AT (Aperio, US), NanoZoomer XR (Hamamatsu, Japan) and NanoZoomer XRL (Hamamatsu, Japan), and includes subjects ranging from 17 to 86 years old. Since the original polygonal annotations do not capture fine-grained normal tissue details between cancerous regions, the dataset was reannotated by expert clinicians at the Norfolk and Norwich University Hospital (NNUH). During this process, 37 slides that presented anomalies or metastases, or that could not be accurately annotated, were excluded. The resultant dataset was partitioned into 100 slides for training and 37 for testing using stratified subset sampling. Slide-level labels were provided for the training and testing sets to denote the subtype present in each slide, and pixel-level segmentations were provided for the test set. These pixel-level annotations were generated using colour-based clustering in Pixelmator Pro \citep{pixelmatorpro2025} and manually refined, leveraging the colour uniformity of H\&E-stained cancerous regions. All annotations were provided by one expert histopathologist and verified by a second. The resultant class distribution is presented in Table \ref{tab:label-distribution}. MBT slides are somewhat under-represented in the test set, since pixel-level annotations could not be reliably provided for some examples after partitioning during final evaluation, thus were removed to avoid skewing results.

\begin{table}[H]
\centering
\begin{tabular}{lrrrrrrrrr}
\toprule
Split & HGSC & LGSC & CCC & EA & MBT & SBT & MA & Normal & Total \\
\midrule
Train      & 35 & 10 & 6 & 8 & 15 & 11 & 4 & 11 & 100 \\
Test & 11  & 4  & 3 & 4 & 3  & 5  & 2 & 5  & 37 \\
\midrule
Total & 46 & 14 & 9 & 12 & 18 & 16 & 6 & 16 & 137 \\
\bottomrule
\end{tabular}
\caption{Slide-level label distribution for the training and test sets.}
\label{tab:label-distribution}
\end{table}

\subsubsection{Hyperparameter Optimisation and Training}
Hyperparameter optimisation was performed independently for all models and ablation experiments. To encompass a wide search space without an exhaustive search, we used Bayesian hyperparameter optimisation with Optuna \citep{optuna}. For each hyperparameter trial, a 5-fold cross validation was performed with 50 epochs, using the macro F1 score of the slide-level predictions for evaluation under class imbalance, and in the absence of instance-level ground truths during training/validation. The hyperparameter search spaces were derived from neighbouring values of the defaults of each method as detailed in their respective papers. All training was performed using weighted sampling to select samples of each subtype with equal frequency, mitigating the bias introduced by the underrepresentation of rarer subtypes. Furthermore, all random seeds for training, weighted sampling, and cross-validation were fixed and consistent for all methods to facilitate reproducibility and fair comparison. The final training was repeated for each model using 5-fold cross validation for 100 epochs. In turn, we arrive at five models that provide a mean and standard deviation for a more robust evaluation. The optimal hyperparameter configuration for MB-DSMIL-CL-PL for our dataset is shown in Table \ref{tab:best_hps}.

\subsubsection{Evaluation}
Instance-level ground truths were derived by assigning each instance the majority class of its constituent pixels, enabling the use of standard classification metrics. Alternative labelling strategies that were considered, such as labelling instances containing any cancerous tissue as cancerous, may introduce label noise when cancerous pixels are sparse, and a threshold-based approach would require the selection of an arbitrary cutoff.

For evaluation on the held-out test set, slide and instance predictions were assessed using both macro F1 and ROC AUC averaged across all five final models of the best hyperparameter combination. For instance-level classification, performance is measured using one-versus-rest (OvR) AUC. For GT versus Normal evaluation, ROC AUC is computed using scores derived from differences in logits between the GT and normal tissue classes, defined as \(s = \hat{y}_{gt} - \hat{y}_{normal}\). In addition, the localisation performance of the attention maps is evaluated using AUC after min-max normalisation to \([0, 1]\).

\subsubsection{Results}

\begin{table}[H]
\centering
\begin{tabular}{lccc}
\toprule
Model &
Slide (Macro F1) &
Instance (Macro F1) &
GT vs. Normal (Macro F1) \\
\midrule
DSMIL                  & 0.672 $\pm$ 0.0926 & 0.301 $\pm$ 0.0329 & 0.567 $\pm$ 0.0584 \\
CLAM                  & 0.757 $\pm$ 0.0703 & 0.309 $\pm$ 0.0701 & 0.376 $\pm$ 0.0784 \\
MB-DSMIL-CL-PL         & \textbf{0.775} $\pm$ 0.0469 & \textbf{0.513} $\pm$ 0.0214 & \textbf{0.684} $\pm$ 0.0297 \\
\midrule
DSMIL-CL-PL* & \textit{0.698} $\pm$ 0.0553 & \textit{0.483} $\pm$ 0.0101 & \textit{0.655} $\pm$ 0.0273 \\
\bottomrule
\end{tabular}
\caption{Comparison of DSMIL, CLAM, and MB-DSMIL-CL-PL (ours) on the slide- and instance-level classification tasks using the mean and standard deviation (mean $\pm$ std) of macro F1 for each model from each fold on the test set. \textit{GT vs. Noraml} indicates that only the slide GT label and normal tissue class logits are considered for the instance classifier's predictions. Bold denotes the best value in each column and italics indicates the second-best. Ablation experiments are denoted by *.}
\label{tab:all_metrics}
\end{table}

\begin{table}[H]
\centering
\resizebox{\linewidth}{!}{
\begin{tabular}{lcccc}
\toprule
Model &
Slide (OvR AUC) &
Instance (OvR AUC) &
GT vs. Normal (AUC) &
Attention (AUC) \\
\midrule
DSMIL                  & \textit{0.952} $\pm$ 0.0196 & 0.781 $\pm$ 0.0373 & 0.580 $\pm$ 0.0534 & 0.756 $\pm$ 0.0405 \\
MB-DSMIL-CL-PL         & \textbf{0.974} $\pm$ 0.00689 & \textbf{0.913} $\pm$ 0.00981 & \textbf{0.807} $\pm$ 0.0242 & \textbf{0.876} $\pm$ 0.00760 \\
\midrule
DSMIL-CL-PL         & 0.949 $\pm$ 0.00988 & \textit{0.873} $\pm$ 0.0103 & \textit{0.782} $\pm$ 0.0171 & \textit{0.854} $\pm$ 0.0126 \\

\bottomrule
\end{tabular}
}
\caption{Comparison of DSMIL, CLAM, and MB-DSMIL-CL-PL (ours) on the slide-classification and localisation tasks using the mean and standard deviation (mean $\pm$ std) of ROC AUC for each model from each fold on the test set. Bold denotes the best value in each column and italics denotes the second-best.}
\label{tab:all_metrics_auc}
\end{table}

\begin{table}[H]
\centering
\begin{tabular}{l cc cc}
\toprule
Class &
\multicolumn{2}{c}{Slide F1} &
\multicolumn{2}{c}{Instance F1} \\
& DSMIL & MB-DSMIL-CL-PL & DSMIL & MB-DSMIL-CL-PL \\
\midrule
MucA & 0.327 $\pm$ 0.306 & \textbf{0.733} $\pm$ 0.389 & 0.183 $\pm$ 0.037 & \textbf{0.417} $\pm$ 0.092 \\
CCC  & \textbf{0.840} $\pm$ 0.080 & 0.620 $\pm$ 0.147 & 0.435 $\pm$ 0.067 & \textbf{0.636} $\pm$ 0.102 \\
EA   & 0.455 $\pm$ 0.376 & \textbf{0.740} $\pm$ 0.111 & 0.174 $\pm$ 0.046 & \textbf{0.392} $\pm$ 0.070 \\
LGSC & 0.583 $\pm$ 0.149 & \textbf{0.590} $\pm$ 0.067 & 0.276 $\pm$ 0.081 & \textbf{0.400} $\pm$ 0.040 \\
SBT  & 0.777 $\pm$ 0.077 & \textbf{0.806} $\pm$ 0.070 & 0.132 $\pm$ 0.054 & \textbf{0.288} $\pm$ 0.018 \\
MBT  & 0.740 $\pm$ 0.177 & \textbf{0.921} $\pm$ 0.102 & 0.262 $\pm$ 0.142 & \textbf{0.456} $\pm$ 0.082 \\
HGSC & \textbf{0.838} $\pm$ 0.088 & 0.832 $\pm$ 0.034 & 0.481 $\pm$ 0.036 & \textbf{0.700} $\pm$ 0.040 \\
N    & 0.820 $\pm$ 0.147 & \textbf{0.956} $\pm$ 0.054 & 0.463 $\pm$ 0.073 & \textbf{0.815} $\pm$ 0.017 \\
\bottomrule
\end{tabular}
\caption{Per-class slide- and instance-level F1 scores on the test set for the model from each fold (mean $\pm$ std). Best performance per class and task is shown in bold.}
\label{tab:per_class_f1}
\end{table}



\begin{figure}[H]
\centering

\newcommand{\mw}{0.32\linewidth}

\parbox[c]{\mw}{\centering GT}
\parbox[c]{\mw}{\centering DSMIL}
\parbox[c]{\mw}{\centering MB-DSMIL-CL-PL}

\vspace{6pt}

\parbox[c]{\mw}{}
\parbox[c]{\mw}{\centering Instance Predictions}
\parbox[c]{\mw}{}

\vspace{4pt}

\begin{subfigure}[H]{\mw}
\centering
\includegraphics[width=\linewidth]{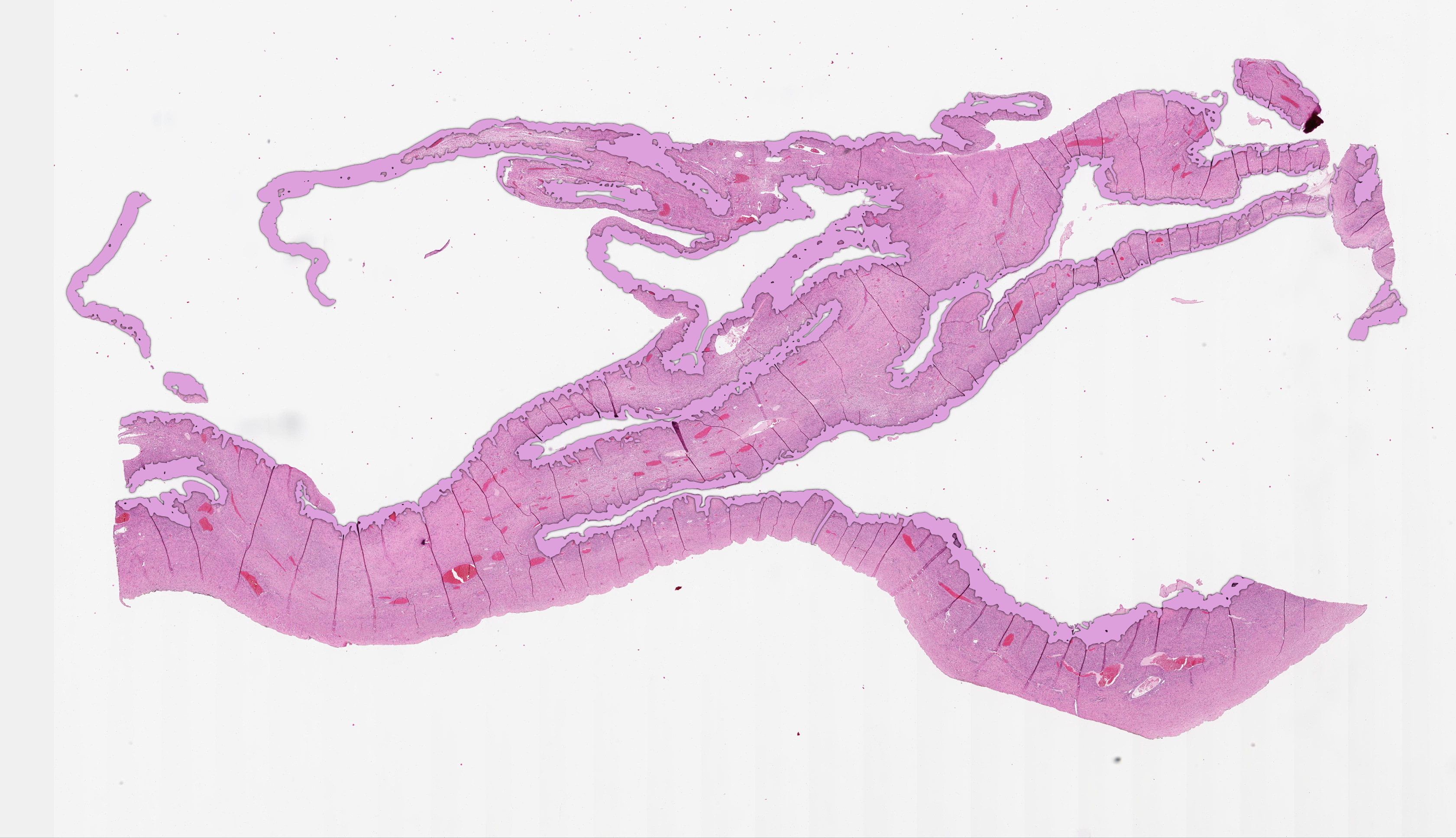}
\end{subfigure}
\begin{subfigure}[H]{\mw}
\centering
\includegraphics[width=\linewidth]{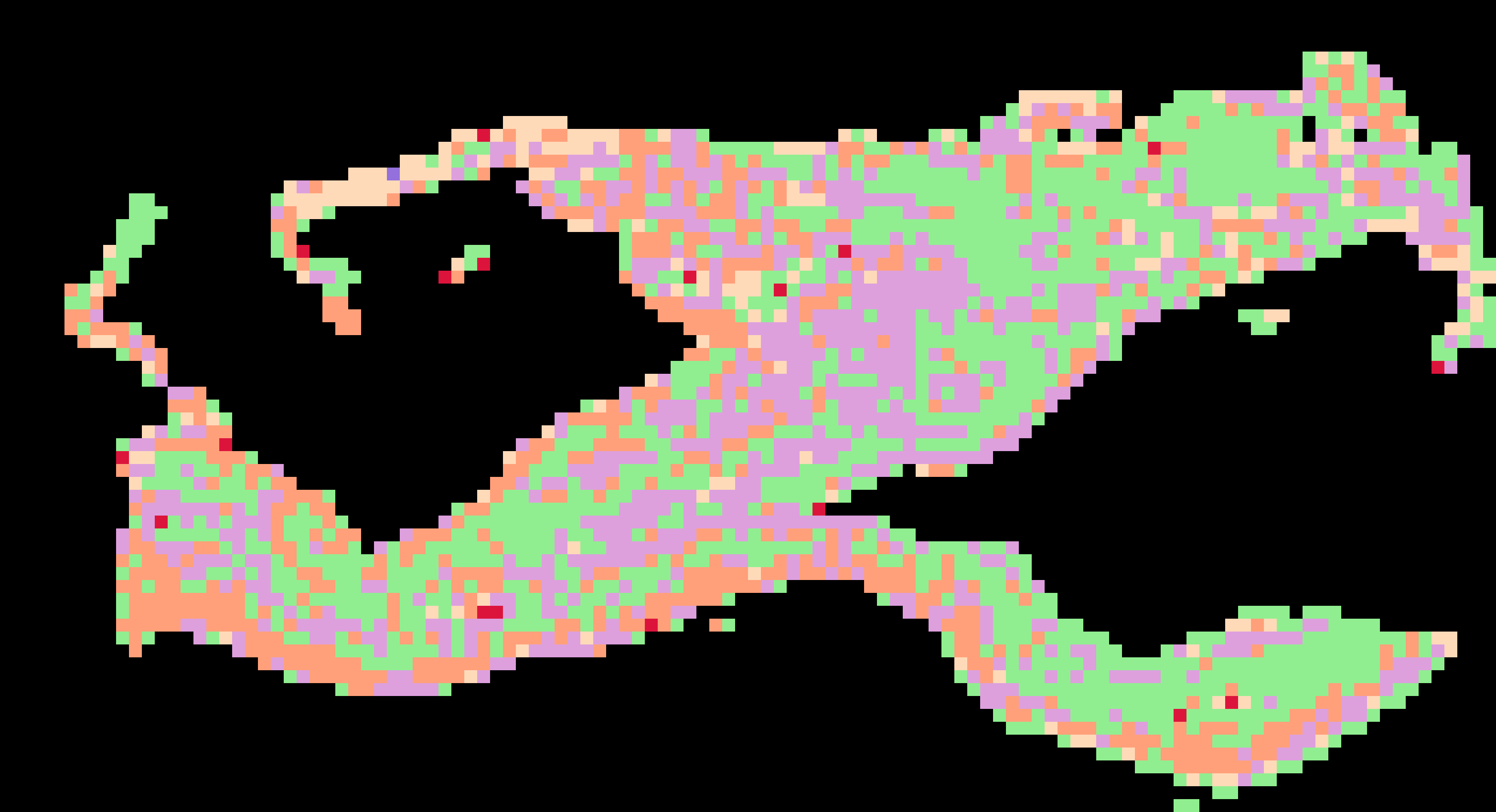}
\end{subfigure}
\begin{subfigure}[H]{\mw}
\centering
\includegraphics[width=\linewidth]{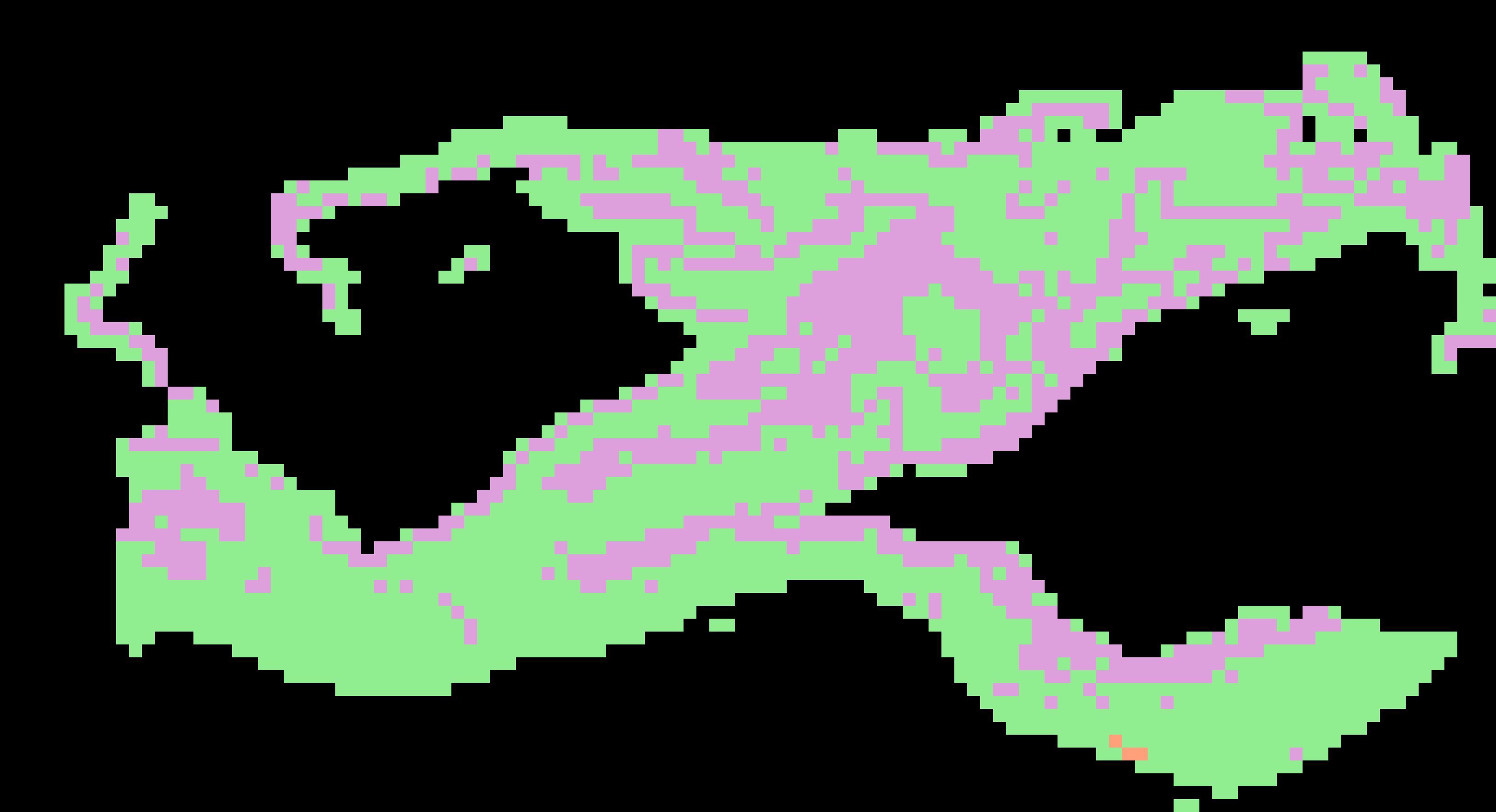}
\end{subfigure}

\vspace{8pt}

\parbox[c]{\mw}{}
\parbox[c]{\mw}{\centering GT vs. Normal}
\parbox[c]{\mw}{}

\vspace{4pt}

\begin{subfigure}[H]{\mw}
\centering
\includegraphics[width=\linewidth]{GT_AI-DROV-144_13748.png}
\end{subfigure}
\begin{subfigure}[H]{\mw}
\centering
\includegraphics[width=\linewidth]{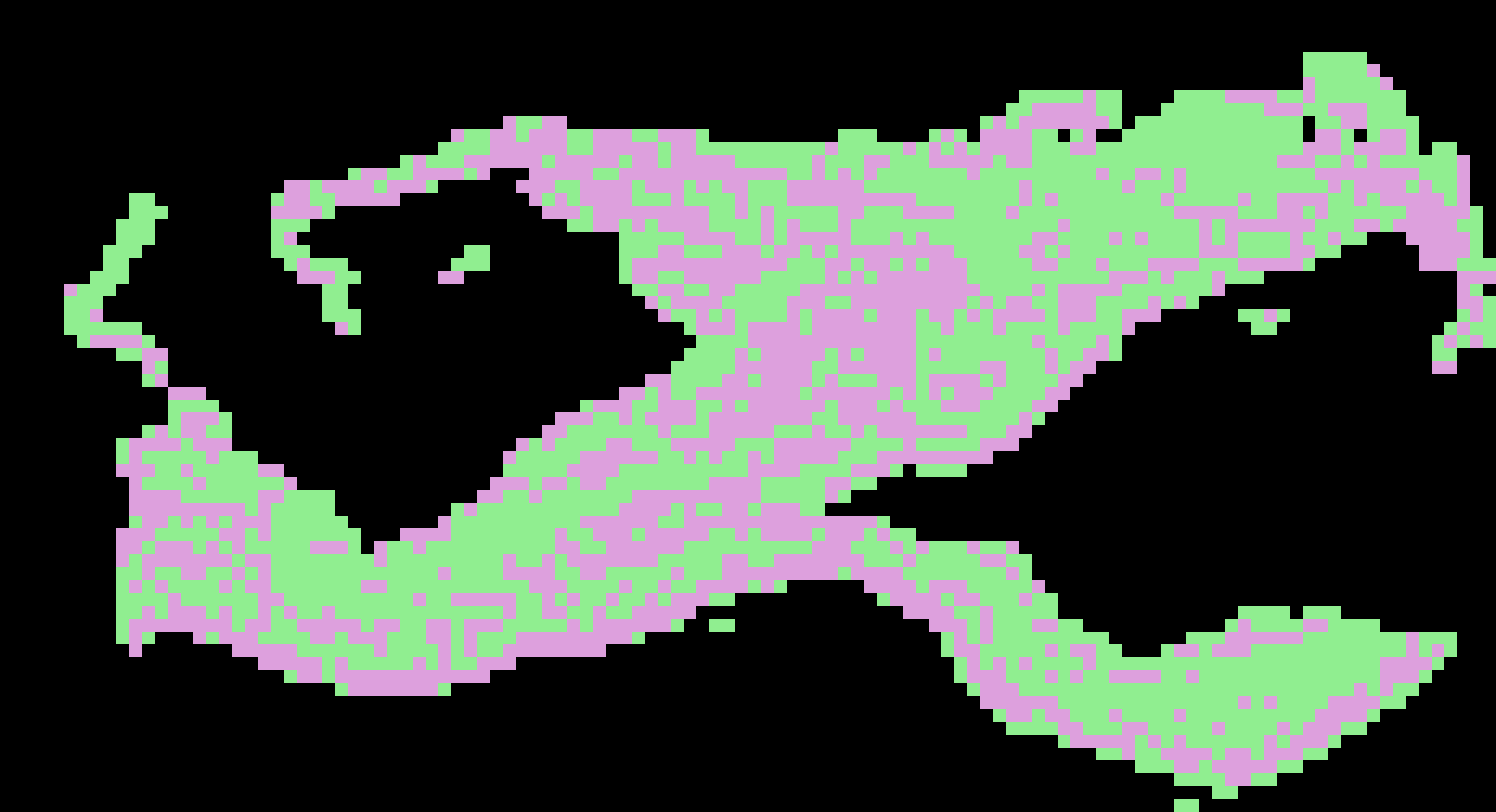}
\end{subfigure}
\begin{subfigure}[H]{\mw}
\centering
\includegraphics[width=\linewidth]{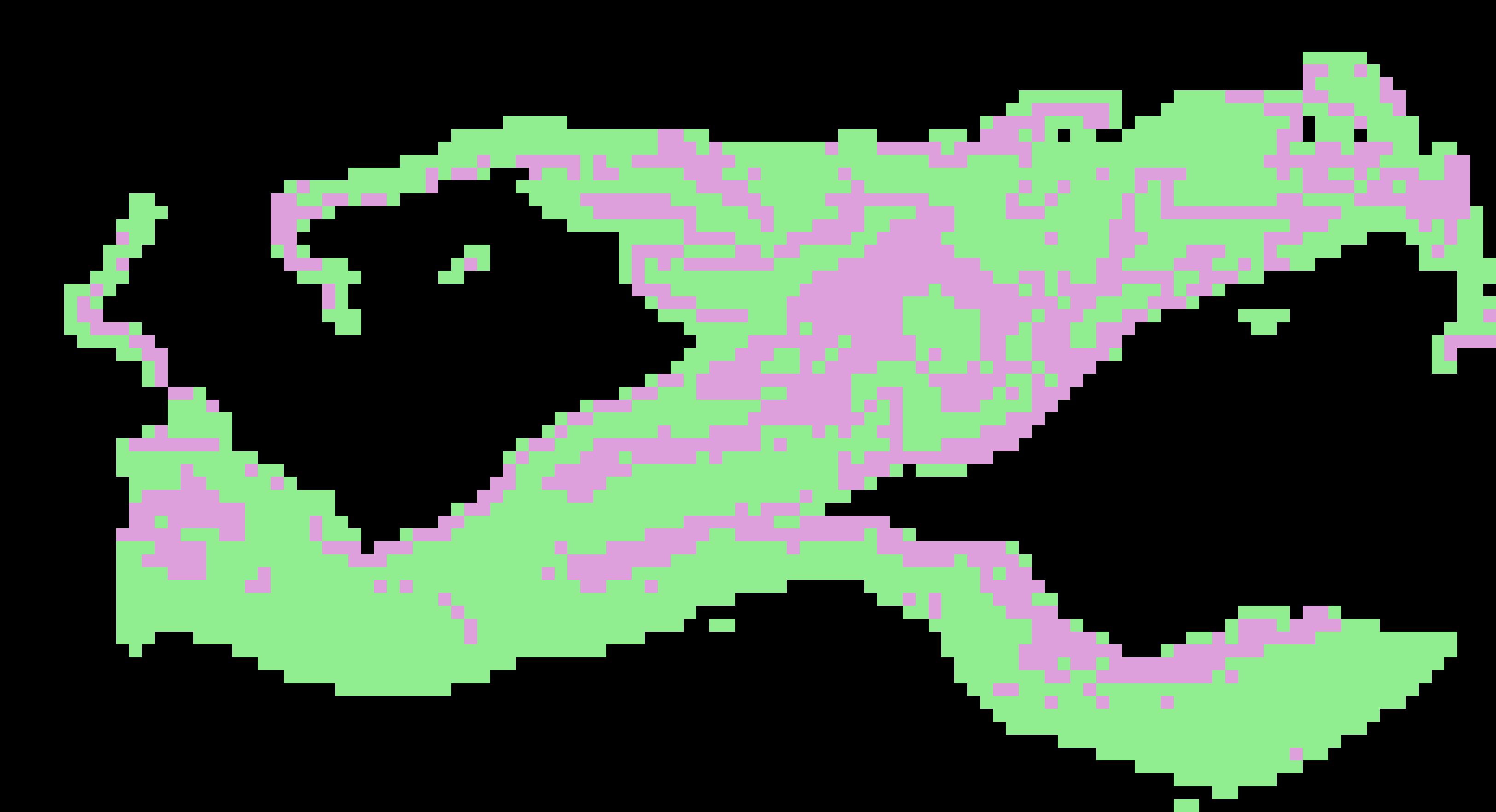}
\end{subfigure}

\vspace{8pt}

\parbox[c]{\mw}{}
\parbox[c]{\mw}{\centering Attention Map}
\parbox[c]{\mw}{}

\vspace{4pt}

\begin{subfigure}[H]{\mw}
\centering
\includegraphics[width=\linewidth]{GT_AI-DROV-144_13748.png}
\end{subfigure}
\begin{subfigure}[H]{\mw}
\centering
\includegraphics[width=\linewidth]{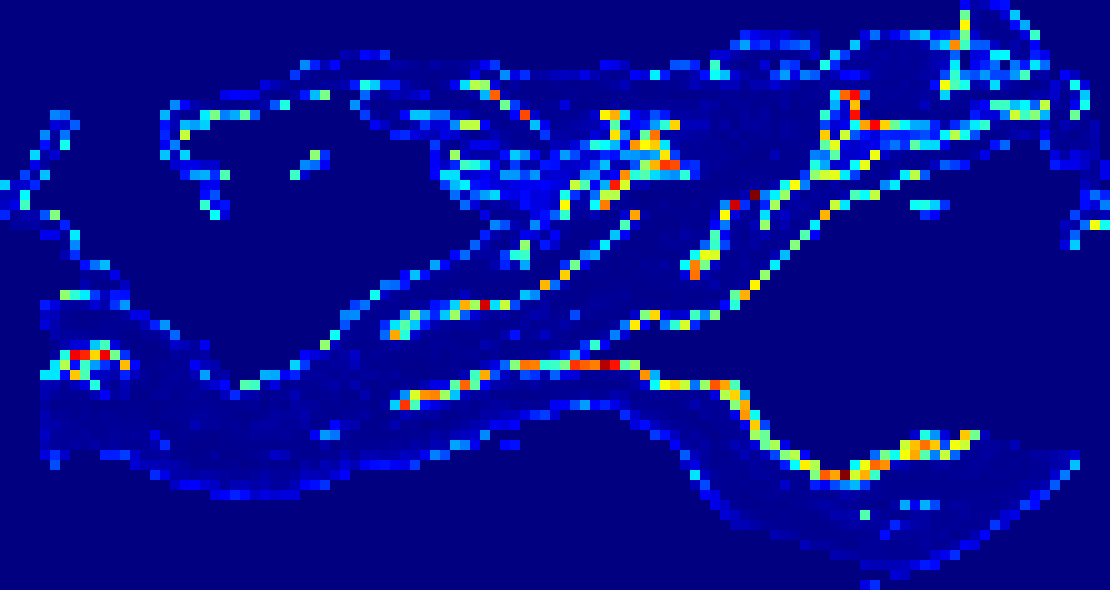}
\end{subfigure}
\begin{subfigure}[H]{\mw}
\centering
\includegraphics[width=\linewidth]{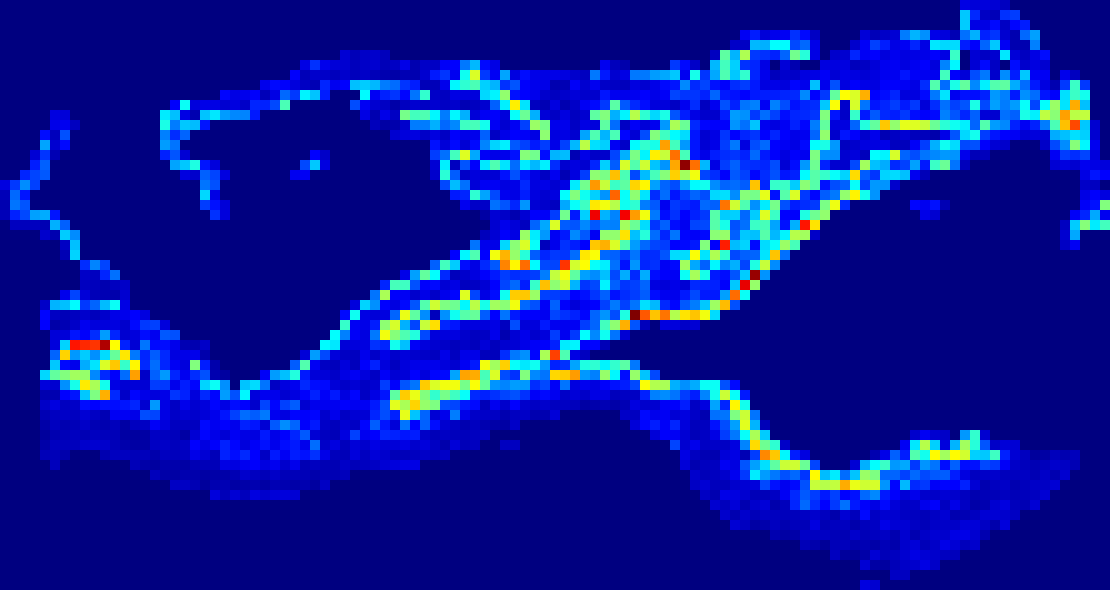}
\end{subfigure}

\caption{Qualitative comparison for an MBT slide (test set). The ground-truth image is identical across methods and shown in the left column for reference. Rows show instance predictions, predictions filtered by GT vs.\ Normal, and attention maps (scaled using Min-Max normalisation).}
\label{fig:qualitative-mbt}
\end{figure}

\begin{figure}[H]
\centering

\newcommand{\mw}{0.32\linewidth}

\parbox[c]{\mw}{\centering GT}
\parbox[c]{\mw}{\centering DSMIL}
\parbox[c]{\mw}{\centering MB-DSMIL-CL-PL}

\vspace{6pt}

\parbox[c]{\mw}{}
\parbox[c]{\mw}{\centering Instance Predictions}
\parbox[c]{\mw}{}

\vspace{4pt}

\begin{subfigure}[H]{\mw}
\centering
\includegraphics[width=\linewidth]{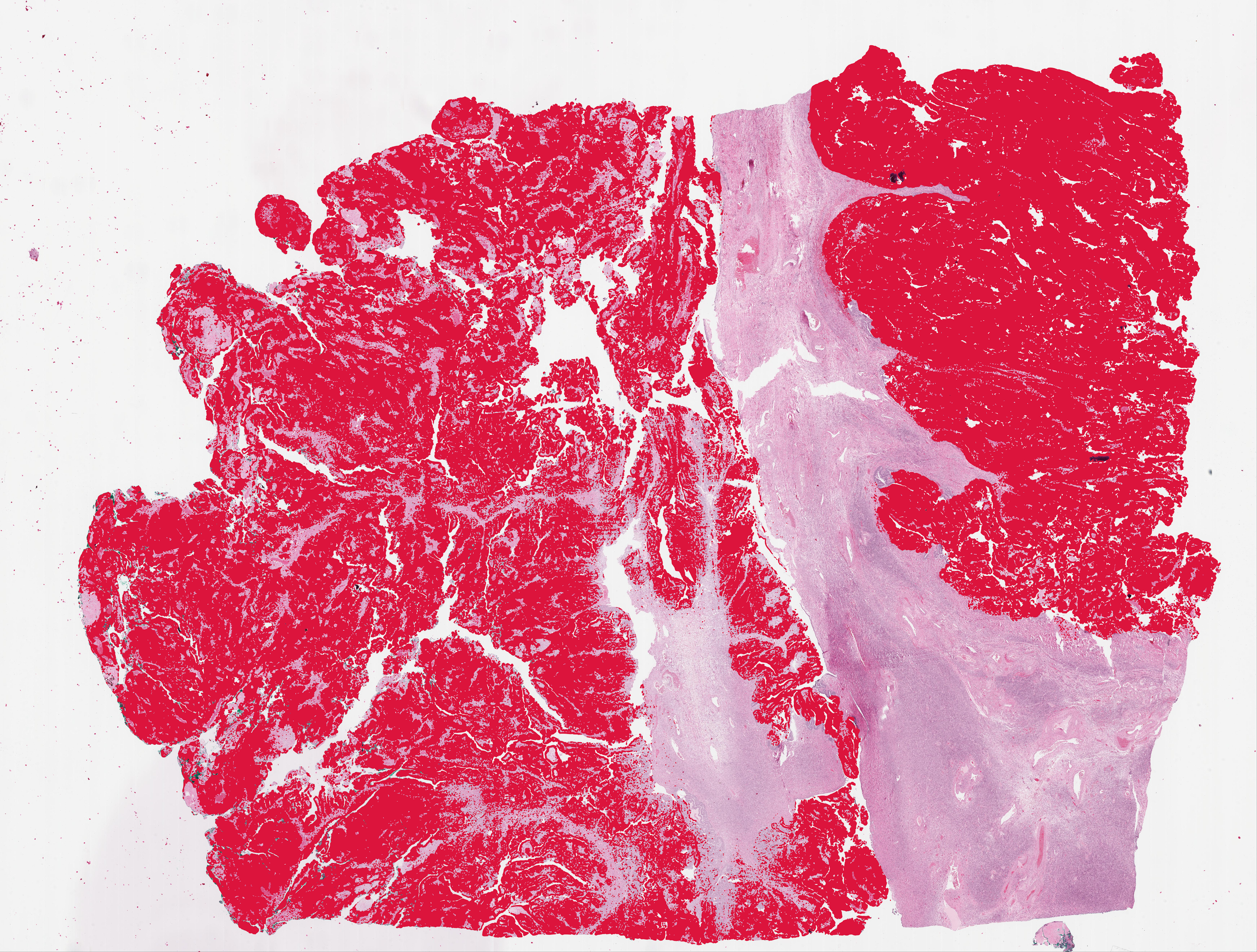}
\end{subfigure}
\begin{subfigure}[H]{\mw}
\centering
\includegraphics[width=\linewidth]{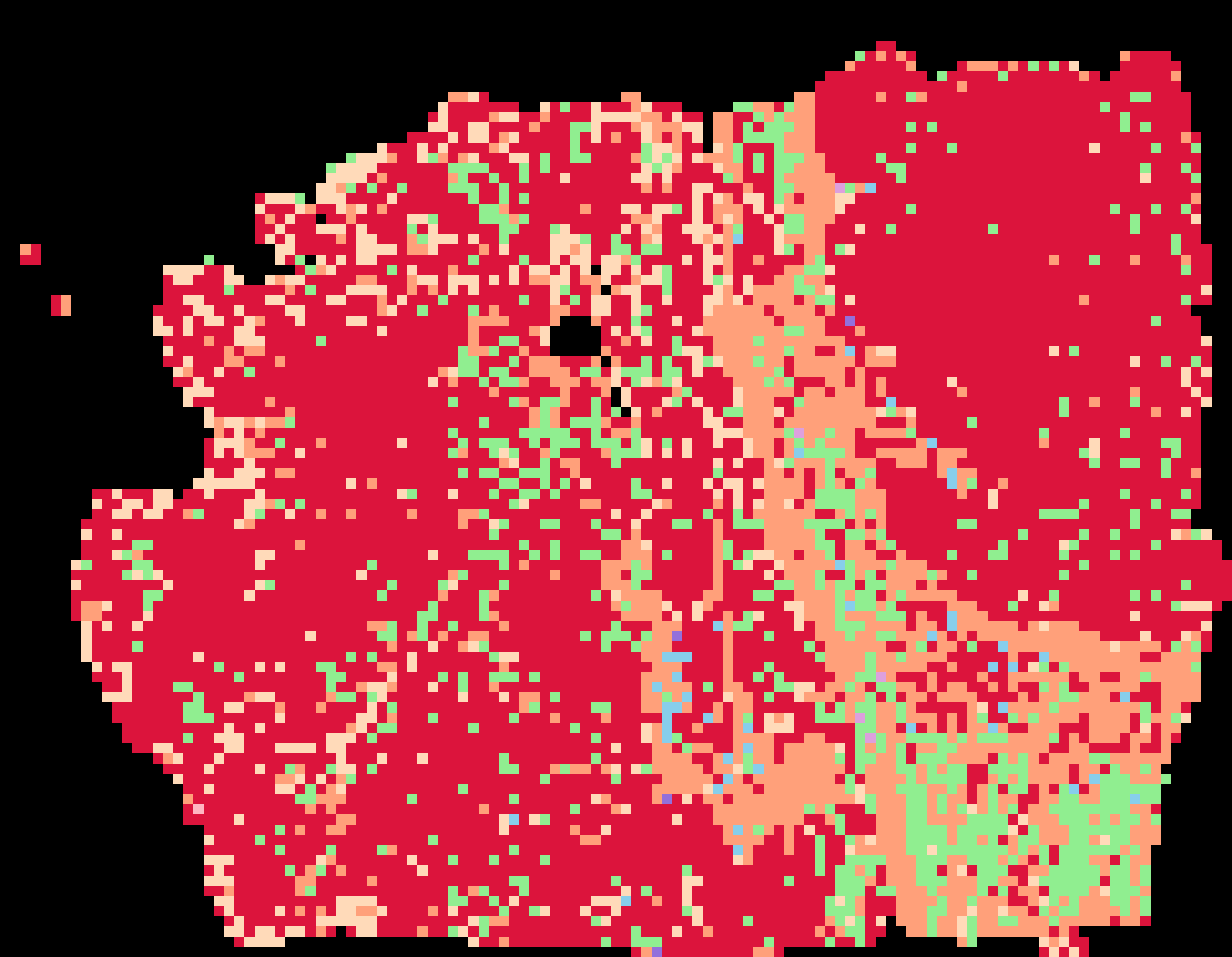}
\end{subfigure}
\begin{subfigure}[H]{\mw}
\centering
\includegraphics[width=\linewidth]{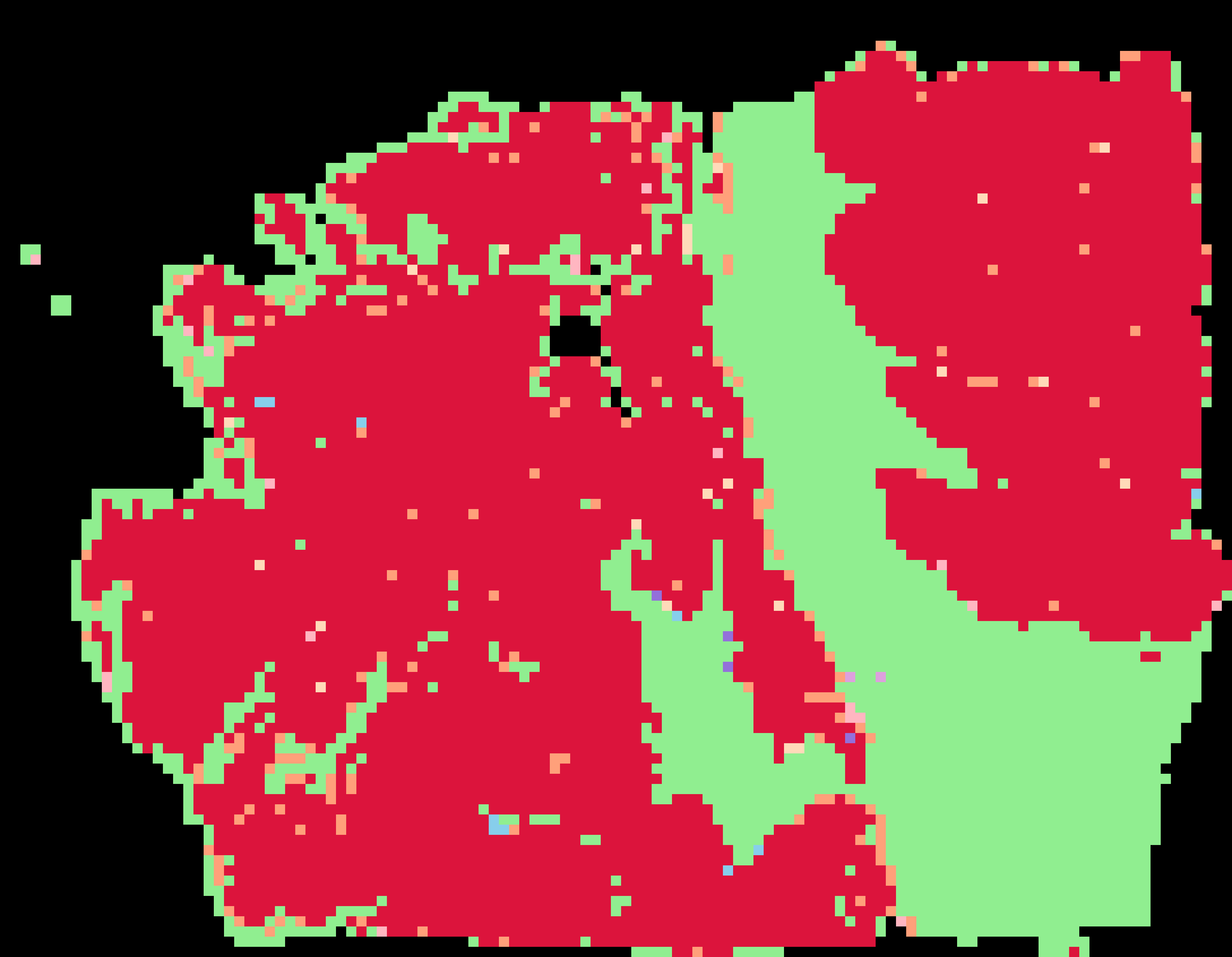}
\end{subfigure}

\vspace{8pt}

\parbox[c]{\mw}{}
\parbox[c]{\mw}{\centering GT vs. Normal}
\parbox[c]{\mw}{}

\vspace{4pt}

\begin{subfigure}[H]{\mw}
\centering
\includegraphics[width=\linewidth]{GT_AI-DROV-037_13002.png}
\end{subfigure}
\begin{subfigure}[H]{\mw}
\centering
\includegraphics[width=\linewidth]{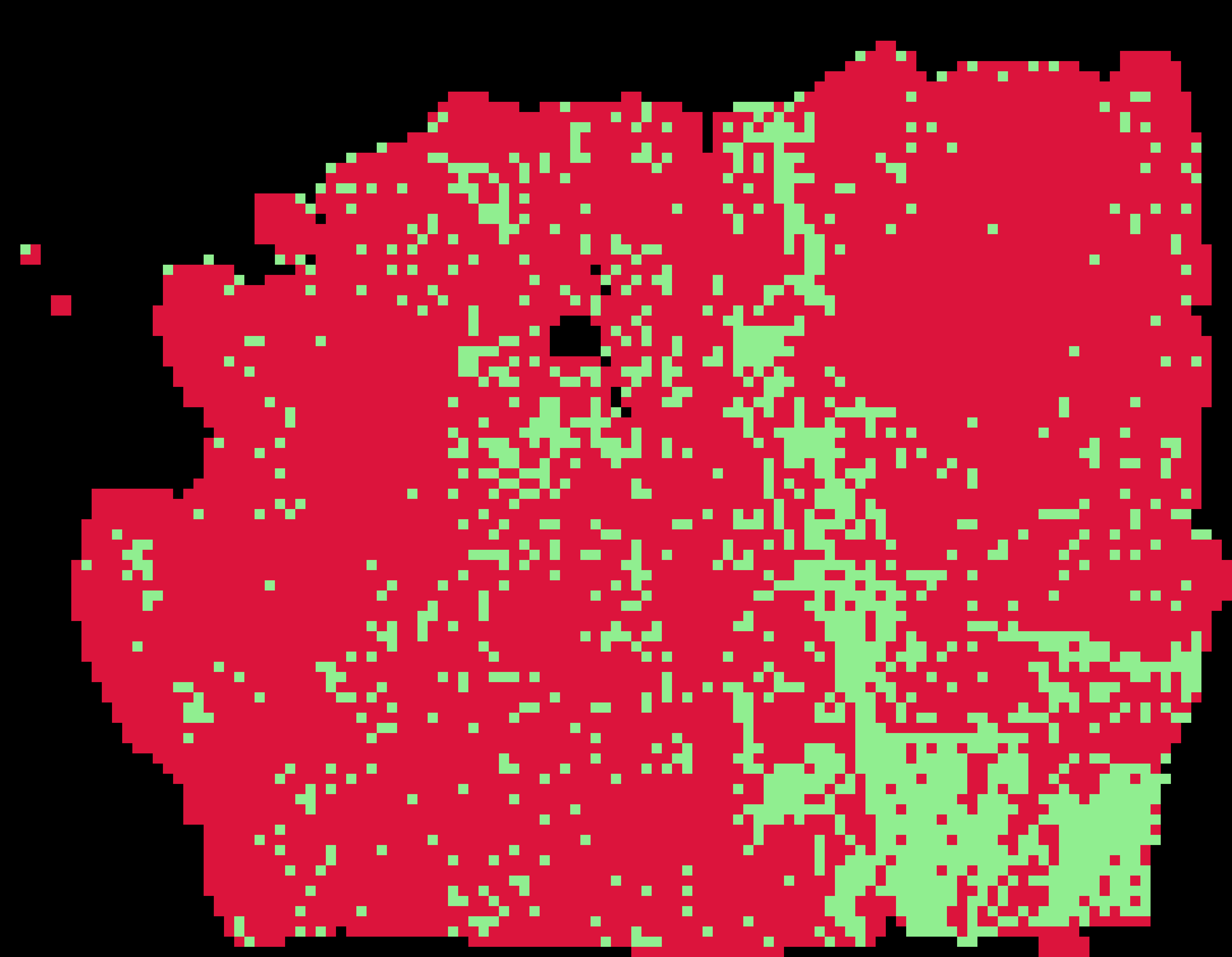}
\end{subfigure}
\begin{subfigure}[H]{\mw}
\centering
\includegraphics[width=\linewidth]{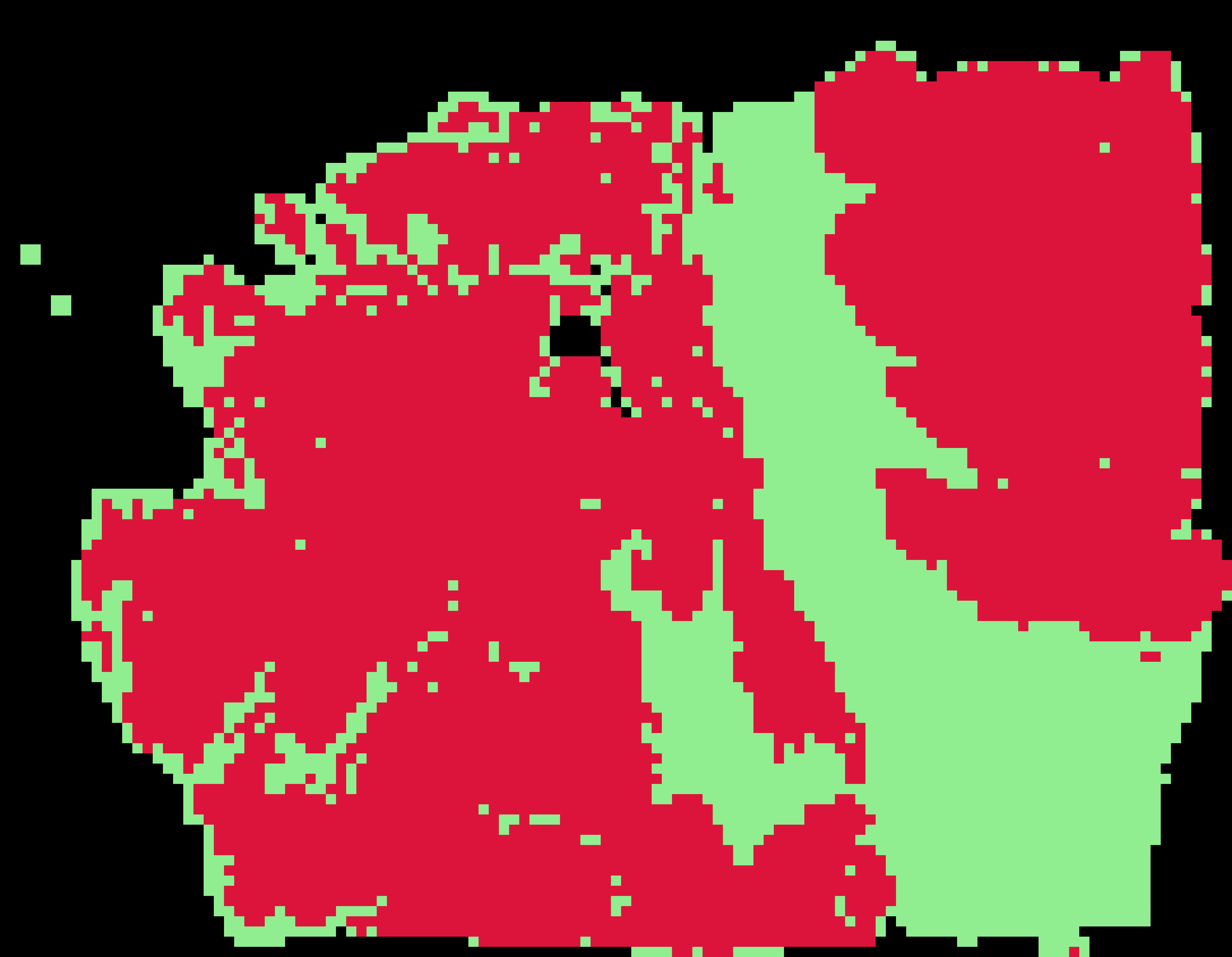}
\end{subfigure}

\vspace{8pt}

\parbox[c]{\mw}{}
\parbox[c]{\mw}{\centering Attention Map}
\parbox[c]{\mw}{}

\vspace{4pt}

\begin{subfigure}[H]{\mw}
\centering
\includegraphics[width=\linewidth]{GT_AI-DROV-037_13002.png}
\end{subfigure}
\begin{subfigure}[H]{\mw}
\centering
\includegraphics[width=\linewidth]{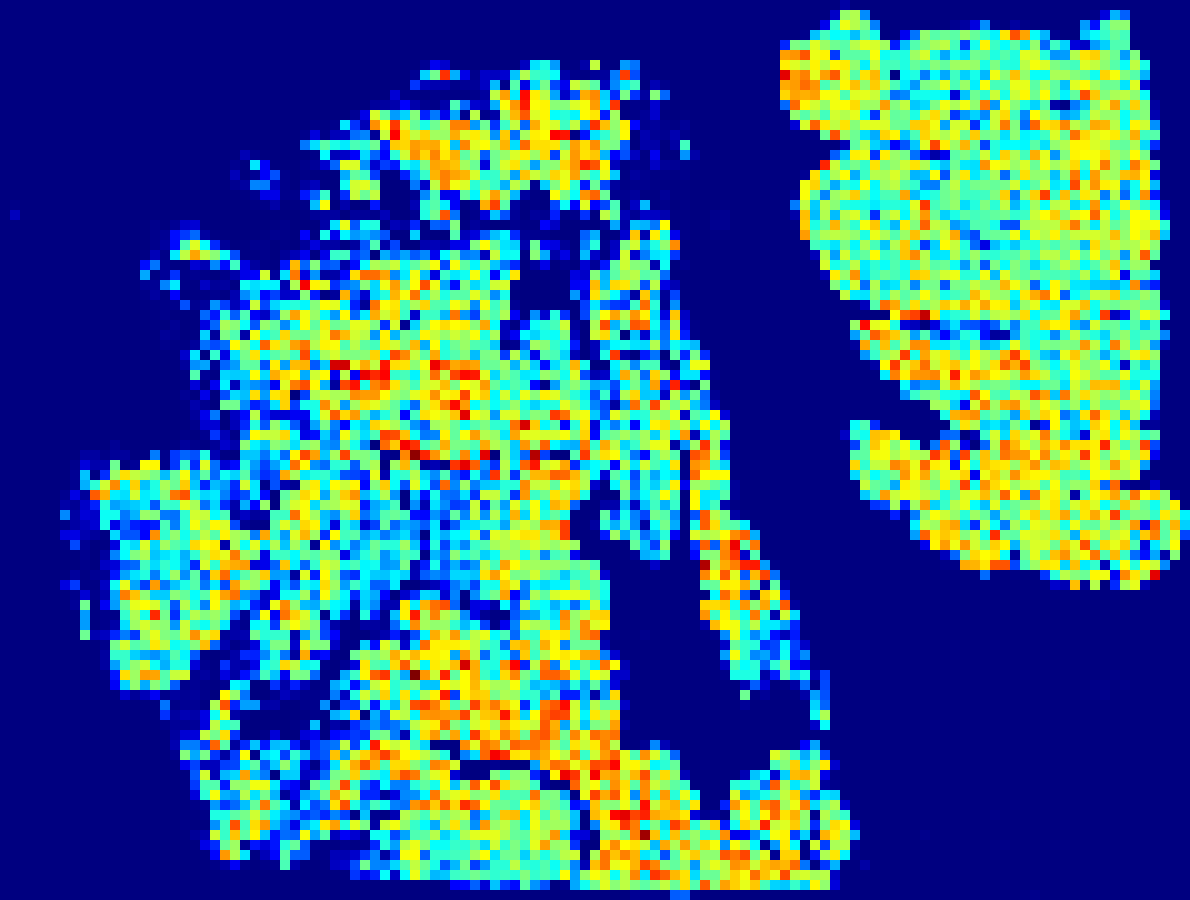}
\end{subfigure}
\begin{subfigure}[H]{\mw}
\centering
\includegraphics[width=\linewidth]{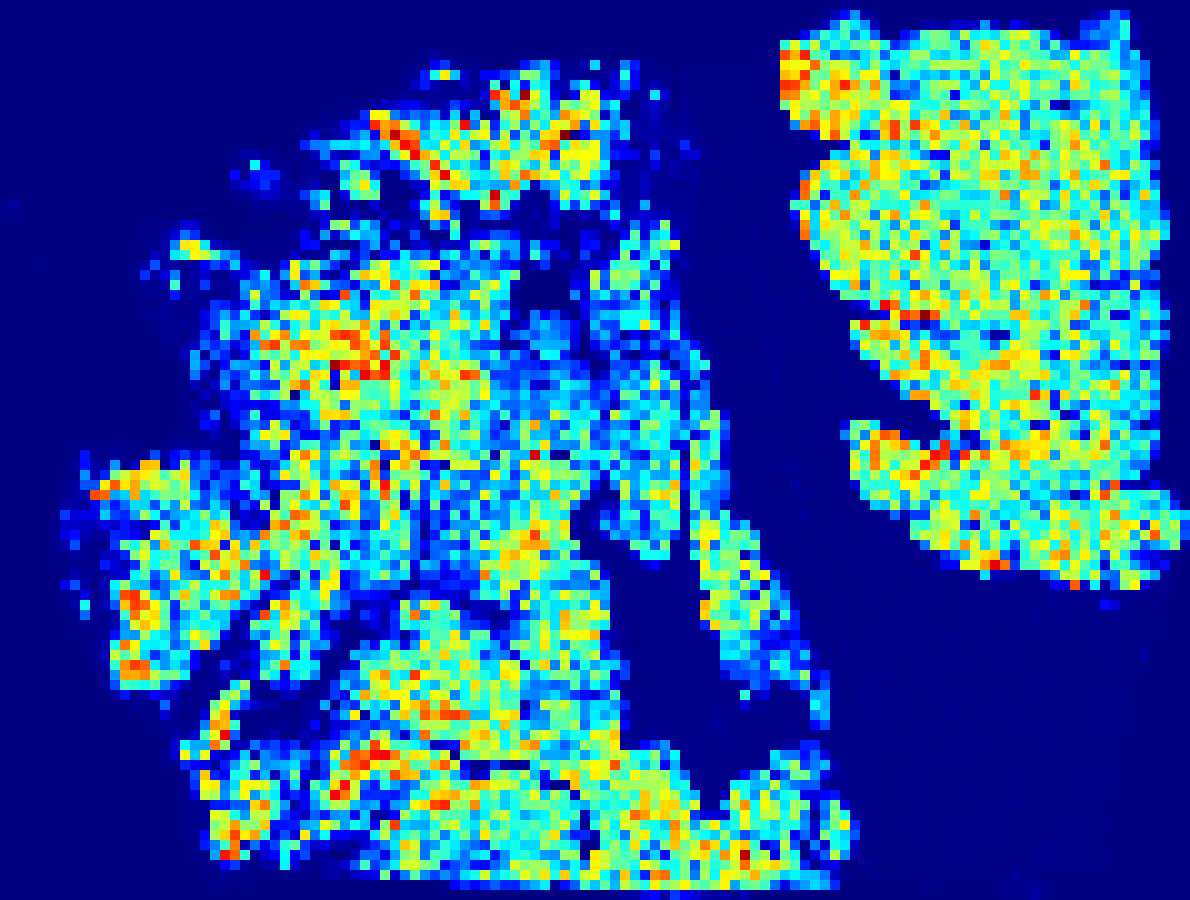}
\end{subfigure}

\caption{Qualitative comparison for a HGSC slide (test set). The ground-truth image is identical across methods and shown in the left column for reference.}
\label{fig:qualitative-hgsc}
\end{figure}

\begin{figure}[H]
\centering

\newcommand{\mw}{0.32\linewidth}

\parbox[c]{\mw}{\centering GT}
\parbox[c]{\mw}{\centering DSMIL}
\parbox[c]{\mw}{\centering MB-DSMIL-CL-PL}

\vspace{6pt}

\parbox[c]{\mw}{}
\parbox[c]{\mw}{\centering Instance Predictions}
\parbox[c]{\mw}{}

\vspace{4pt}

\begin{subfigure}[H]{\mw}
\centering
\includegraphics[width=\linewidth]{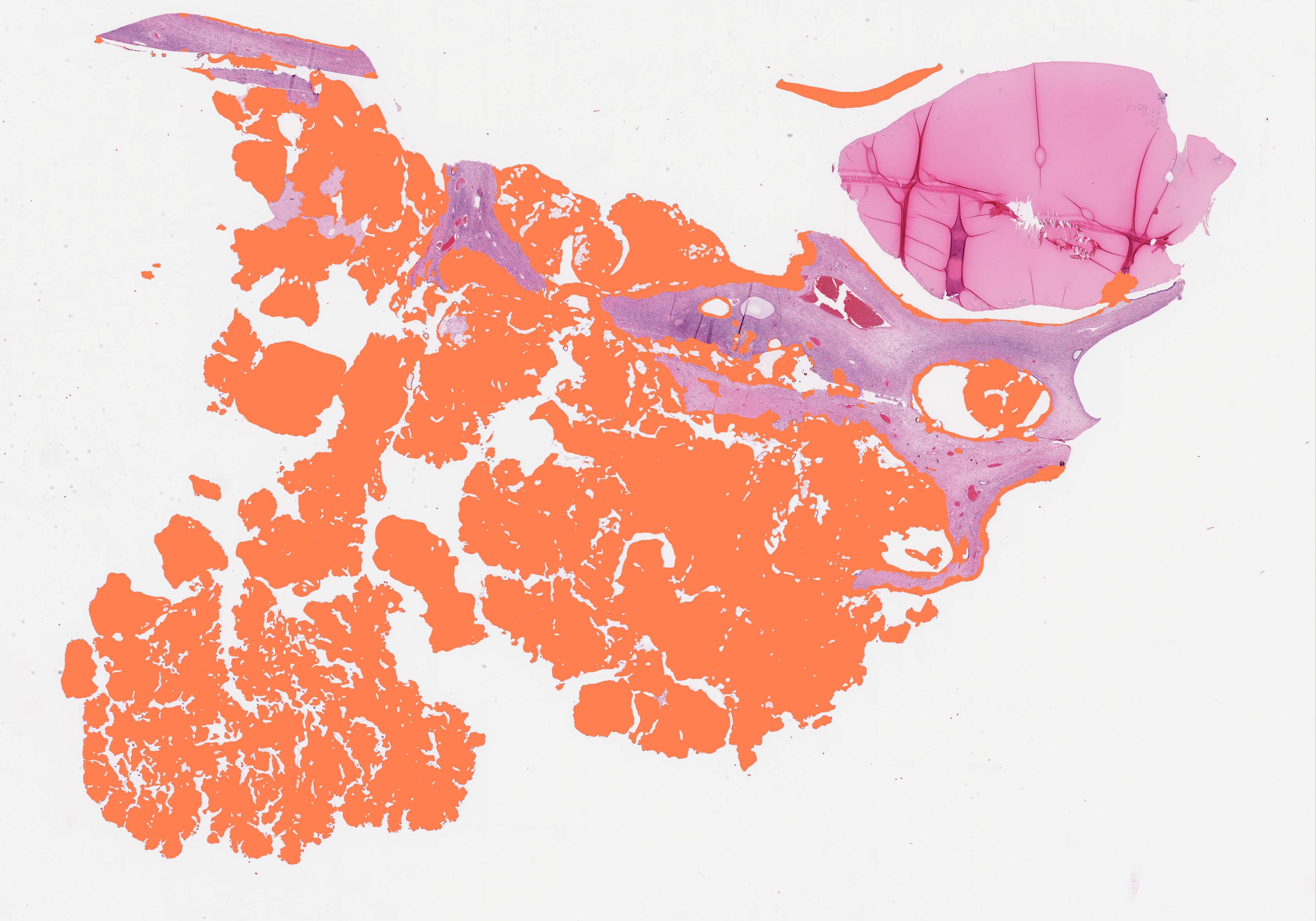}
\end{subfigure}
\begin{subfigure}[H]{\mw}
\centering
\includegraphics[width=\linewidth]{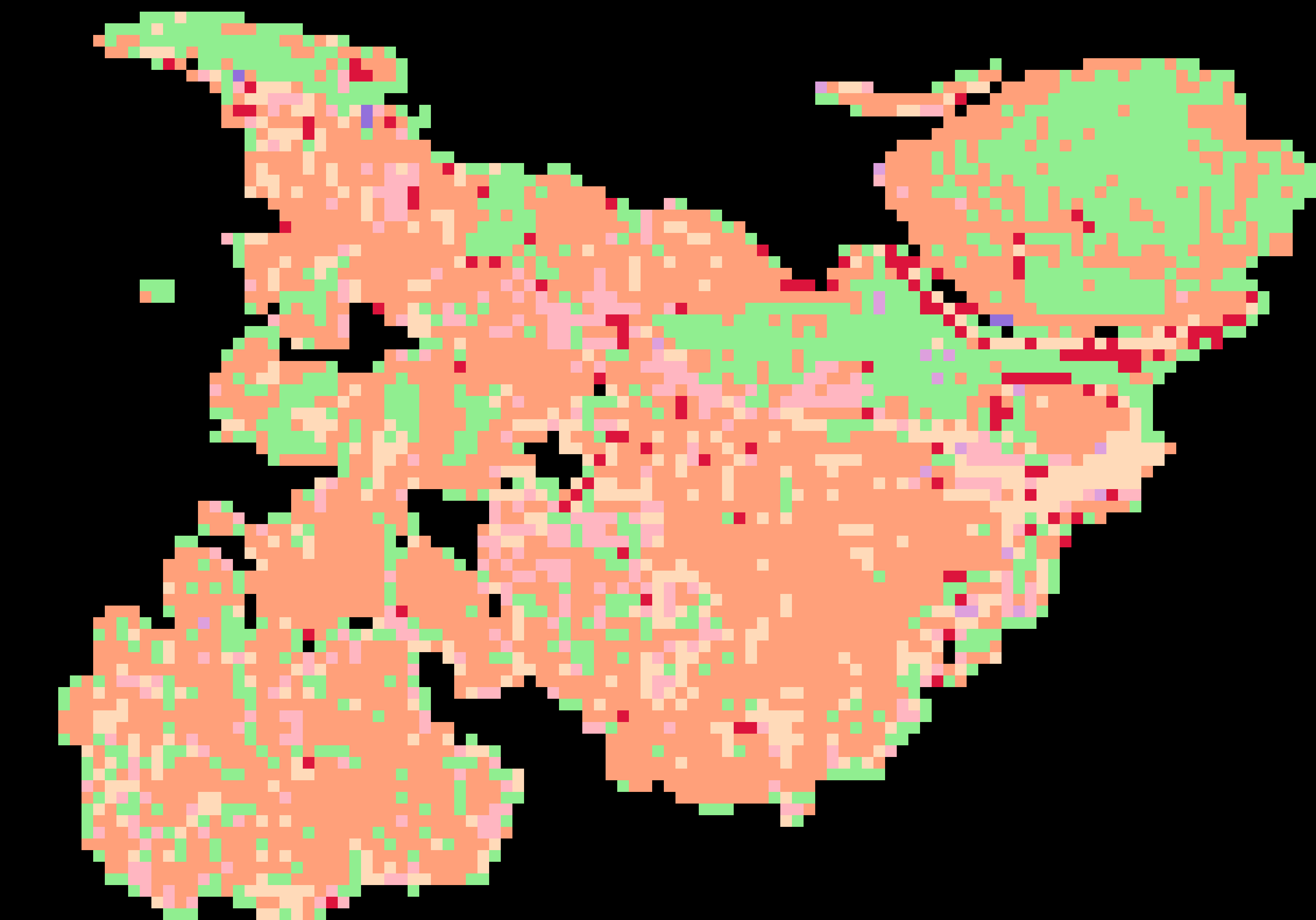}
\end{subfigure}
\begin{subfigure}[H]{\mw}
\centering
\includegraphics[width=\linewidth]{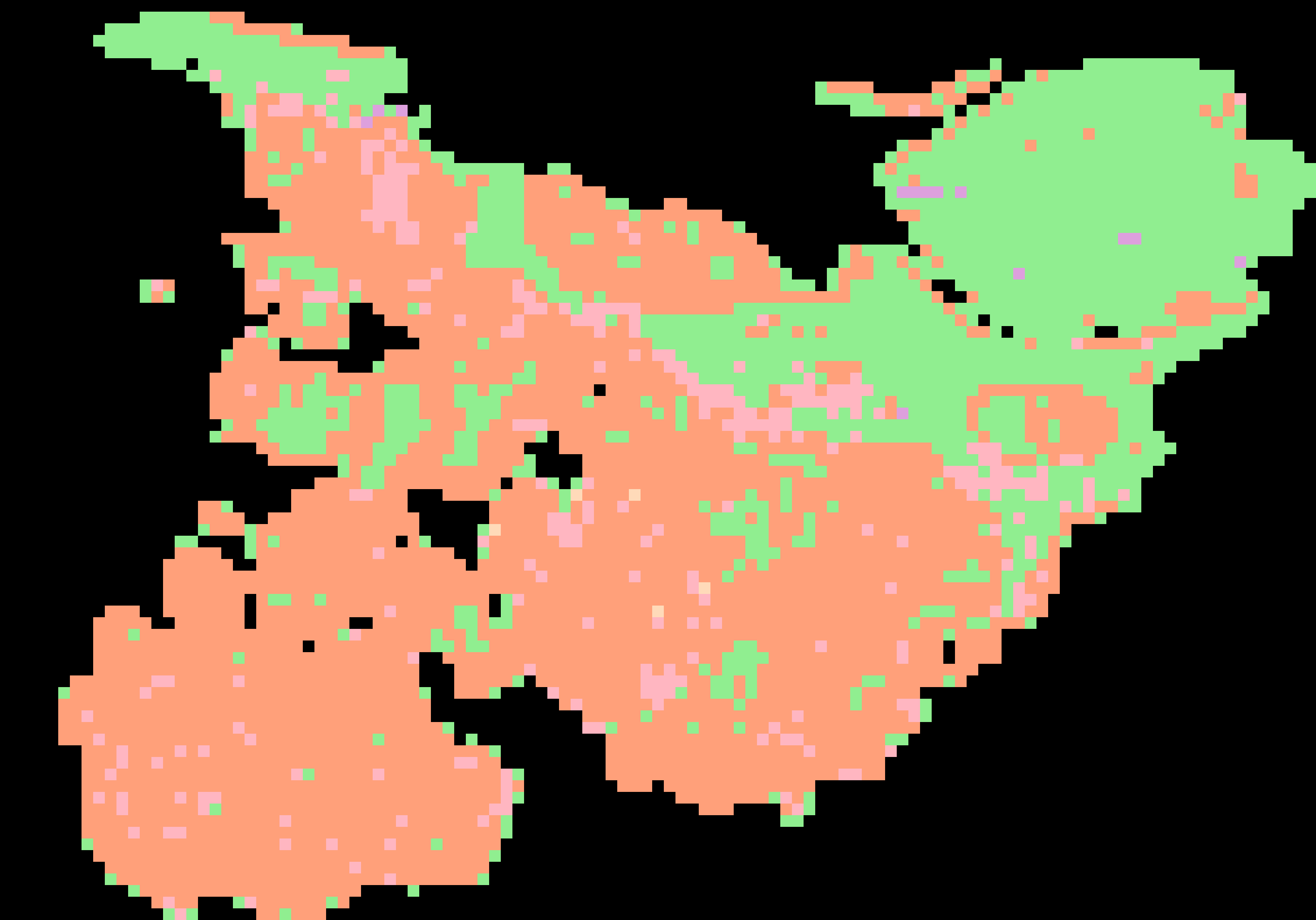}
\end{subfigure}

\vspace{8pt}

\parbox[c]{\mw}{}
\parbox[c]{\mw}{\centering GT vs. Normal}
\parbox[c]{\mw}{}

\vspace{4pt}

\begin{subfigure}[H]{\mw}
\centering
\includegraphics[width=\linewidth]{GT_AI-DROV-090_13060.png}
\end{subfigure}
\begin{subfigure}[H]{\mw}
\centering
\includegraphics[width=\linewidth]{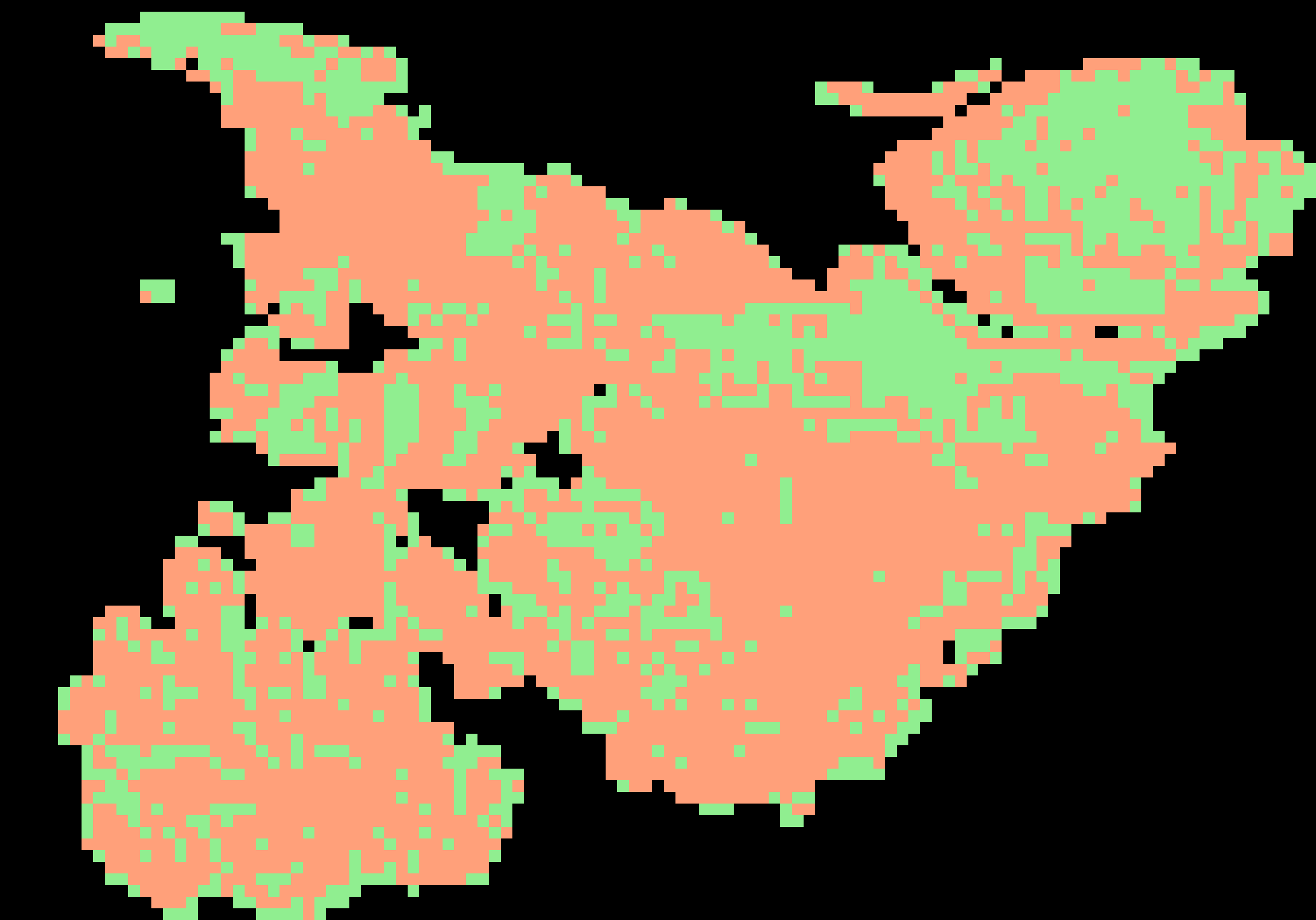}
\end{subfigure}
\begin{subfigure}[H]{\mw}
\centering
\includegraphics[width=\linewidth]{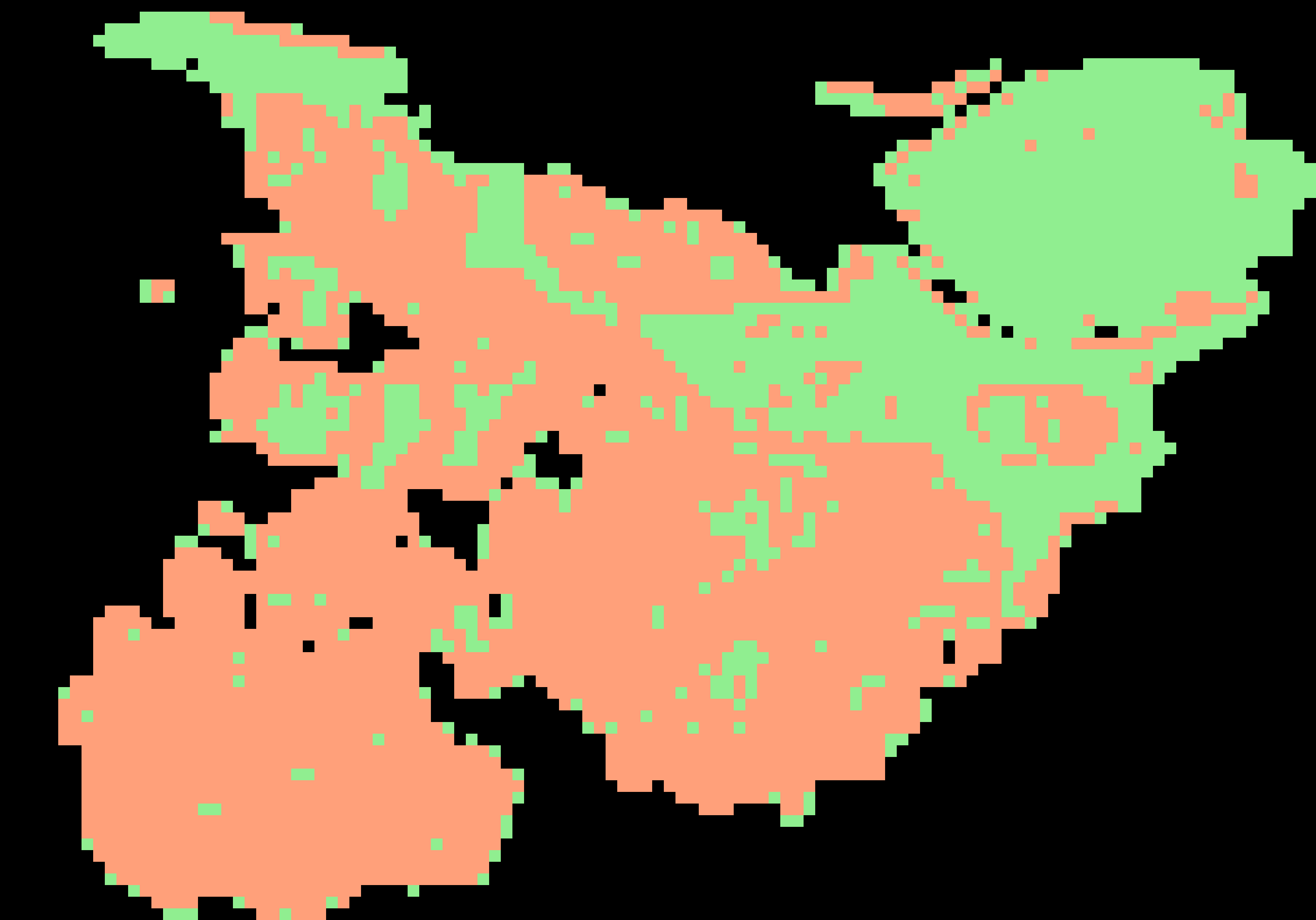}
\end{subfigure}

\vspace{8pt}

\parbox[c]{\mw}{}
\parbox[c]{\mw}{\centering Attention Map}
\parbox[c]{\mw}{}

\vspace{4pt}

\begin{subfigure}[H]{\mw}
\centering
\includegraphics[width=\linewidth]{GT_AI-DROV-090_13060.png}
\end{subfigure}
\begin{subfigure}[H]{\mw}
\centering
\includegraphics[width=\linewidth]{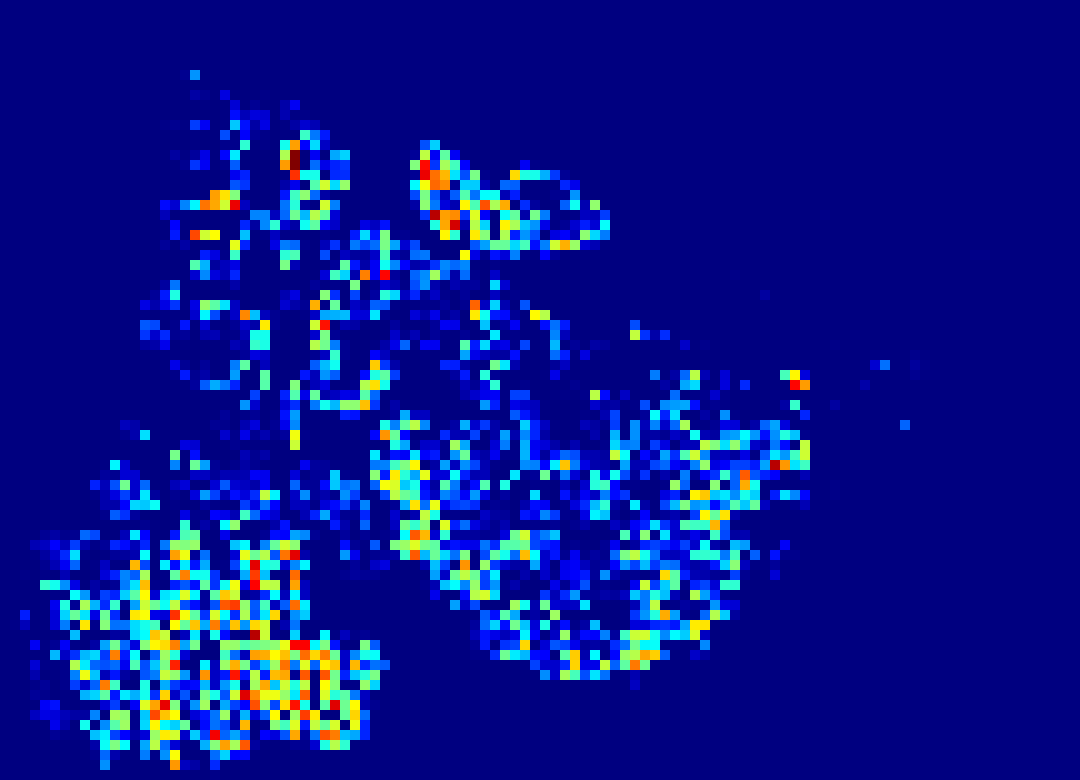}
\end{subfigure}
\begin{subfigure}[H]{\mw}
\centering
\includegraphics[width=\linewidth]{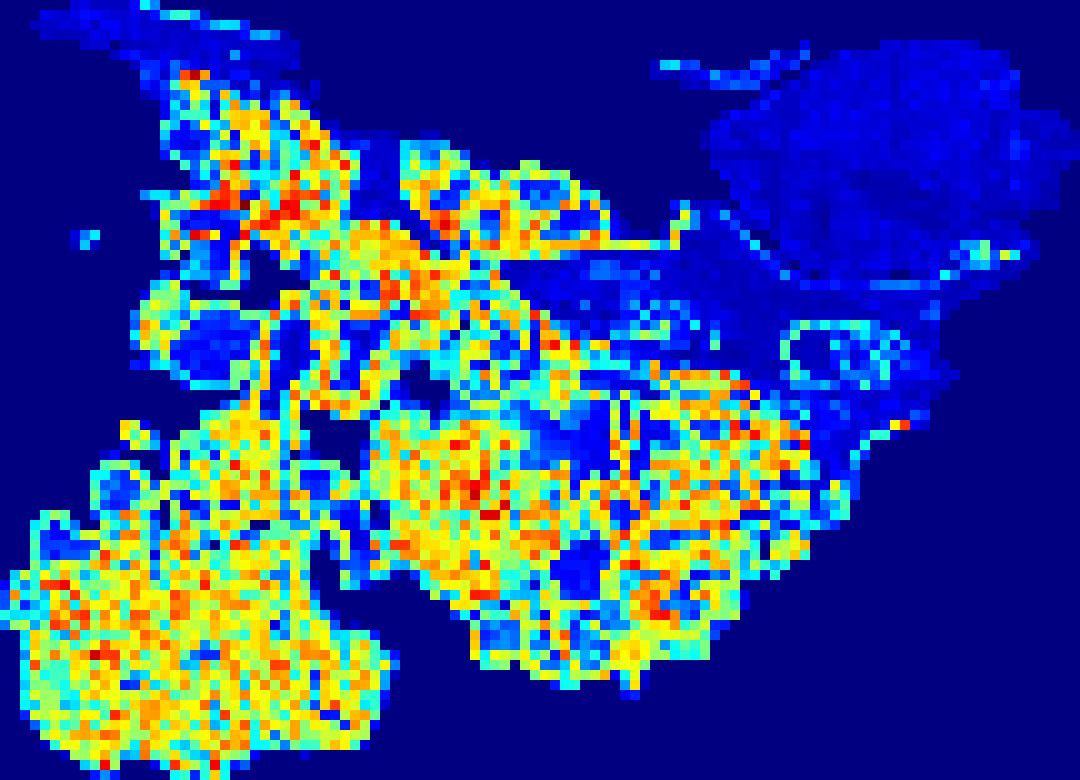}
\end{subfigure}

\caption{Qualitative comparison for an SBT slide (test set). The ground-truth image is identical across methods and shown in the left column for reference.}
\label{fig:qualitative-sbt}
\end{figure}

Tables \ref{tab:all_metrics}, \ref{tab:all_metrics_auc} and \ref{tab:per_class_f1} demonstrate that MB-DSMIL-CL-PL significantly outperforms DSMIL at both the slide and instance levels, with much higher consistency across classes at the slide-level, particularly for minority classes. We also observe a gain in instance classification performance of 16.9\% and 70.4\% in AUC and macro F1, respectively.

Figures \ref{fig:qualitative-mbt}-\ref{fig:qualitative-sbt} present qualitative comparisons of the instance classifier between DSMIL and MB-DSMIL-CL-PL. In all examples, we observe a significant reduction in confusion both between subtypes and between slide-level ground truth and normal tissue after filtering; this is further supported by the confusion matrices in Figure \ref{fig:confusion_matrices}. Moreover, the attention maps produced by MB-DSMIL-CL-PL exhibit reduced sparsity, leading to improved localisation performance via attention. Additionally, we find that these attention maps are more closely aligned with the corresponding instance-level classifications. We hypothesise that this enhanced attention scoring and instance classifier alignment is attributed to supervised contrastive learning, together with class-specific query projection layers in MB-DSMIL. Supervised contrastive learning promotes the alignment of instance embeddings within the same class while increasing inter-class separation, and the class-specific query projection layers further enhance discriminative capacity when determining class relevance within the attention mechanism.

Furthermore, these findings help explain the observed improvement in slide-level classification performance, as enhanced attention scoring provides a more discriminative bag-level feature representation after attention-based aggregation. In addition, incorporating a broader set of relevant instance features has been shown to improve the generalisation performance of bag classifiers in MIL \citep{WENO}, as the bag classifier is less likely to overfit to fewer, easily distinguishable instances.

\section{Conclusion and Future Work} \label{sec:conclusion}
In this work, we proposed a novel MIL architecture for the classification and localisation of histological subtypes of ovarian carcinoma, using only slide-level labels and precomputed features. Our novel approach MB-DSMIL-CL-PL offers a significant improvement of 16.9\% in instance-level AUC and 70.4\% in macro F1 over DSMIL for multi-class settings by introducing the task-specific adaptability of end-to-end contrastive learning and the stability of class prototype learning, while preserving the high computational efficiency and scalability provided by training on precomputed features.

We envisage that future work could incorporate the multi-resolution framework described in \citep{DSMIL} to offer greater performance. In addition, we show that our approach reduces the reliance on a small subset of instances for slide classification, offering greater generalisation. However, additional approaches could be used to maximise this effect, such as random instance dropout \citep{breen_multi_resolution} or the hard-positive instance mining strategy of \cite{WENO} to randomly drop instances of high confidence. Moreover, application to binary classification settings for benchmark datasets such as CAMELYON \citep{camelyon} could easily be achieved using only class prototypes for normal and cancerous tissue, and using binary cross entropy loss for slide level classification.


\appendix
\section{Supplementary Figures}
\label{app1}

\begin{table}[H]
\centering
\begin{tabular}{lr}
\toprule
Hyperparameter & Value \\
\midrule
Learning rate        & 0.0001 \\
Weight decay                & 0.001 \\
Query embedding dim ($q$) & 64 \\
Feature size ($e$)                & 1024 \\
$\eta_w$                    & 0.05 \\
$\eta_s$                    & 0.4 \\
Weak dropout ($p_w$)                 & 0.0 \\
Strong dropout ($p_w$)                & 0.0 \\
\bottomrule
\end{tabular}
\caption{Optimal hyperparameter configuration for MB-DSMIL-CL-PL for our dataset.}
\label{tab:best_hps}
\end{table}

\begin{figure}[H]
\begin{subfigure}[H]{0.49\textwidth}
\centering
\includegraphics[width=\textwidth]{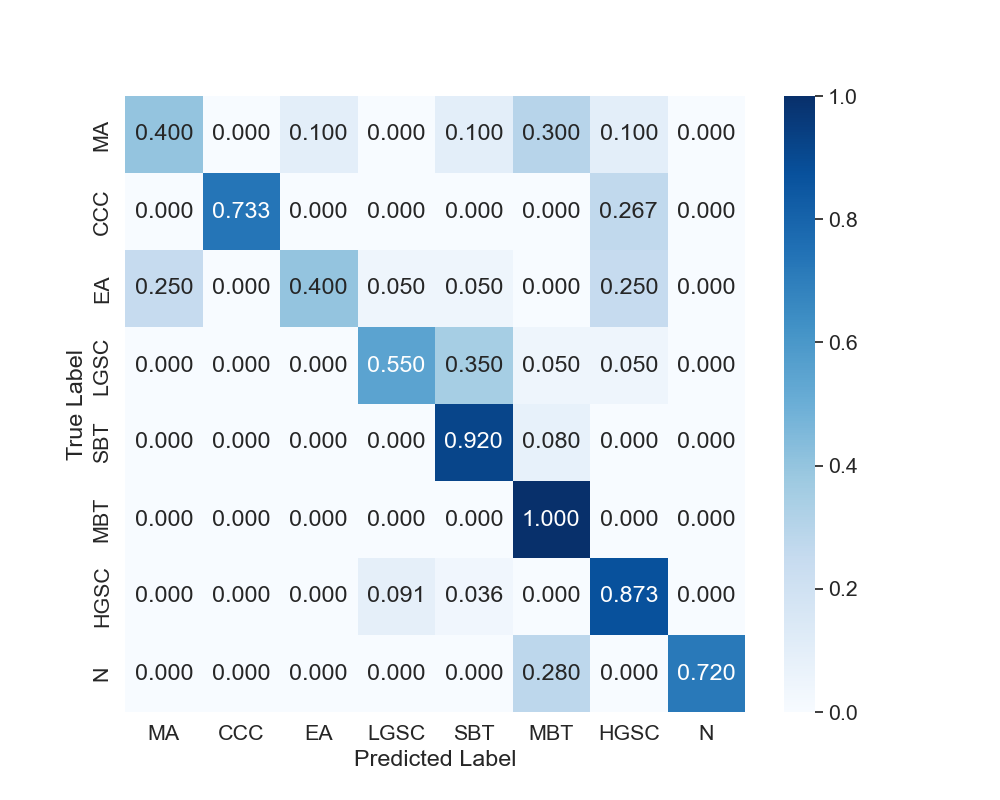}
\caption{}
\end{subfigure}
\begin{subfigure}[H]{0.49\textwidth}
\centering
\includegraphics[width=\textwidth]{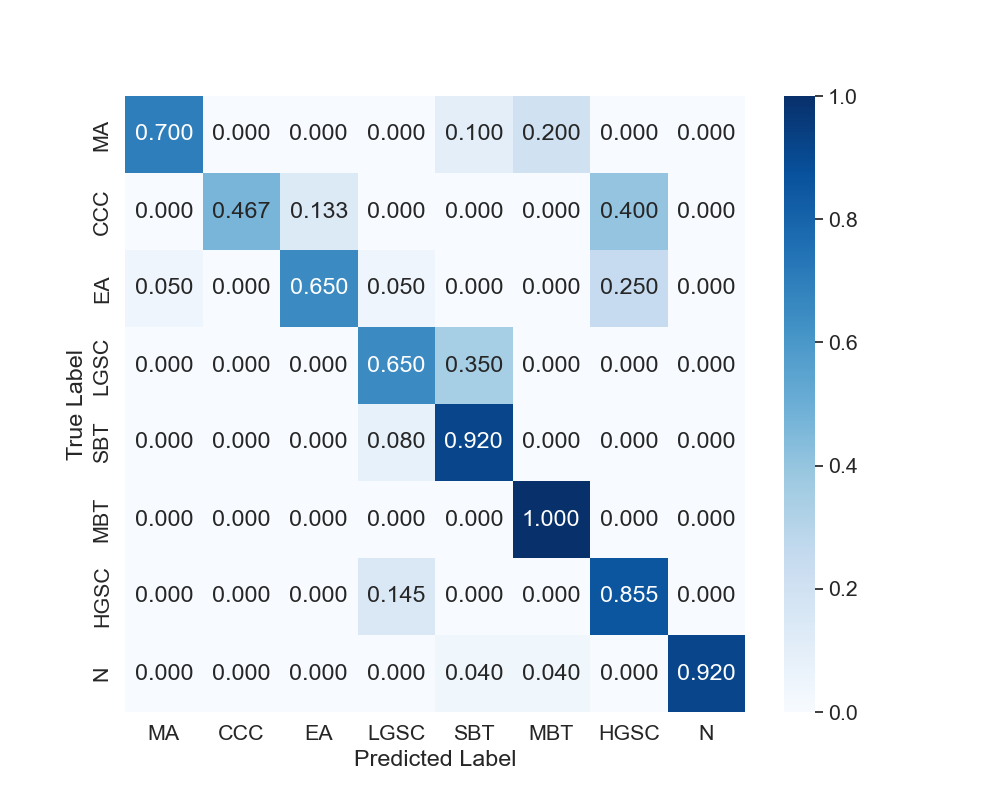}
\caption{}
\end{subfigure}

\begin{subfigure}[H]{0.49\textwidth}
\centering
\includegraphics[width=\textwidth]{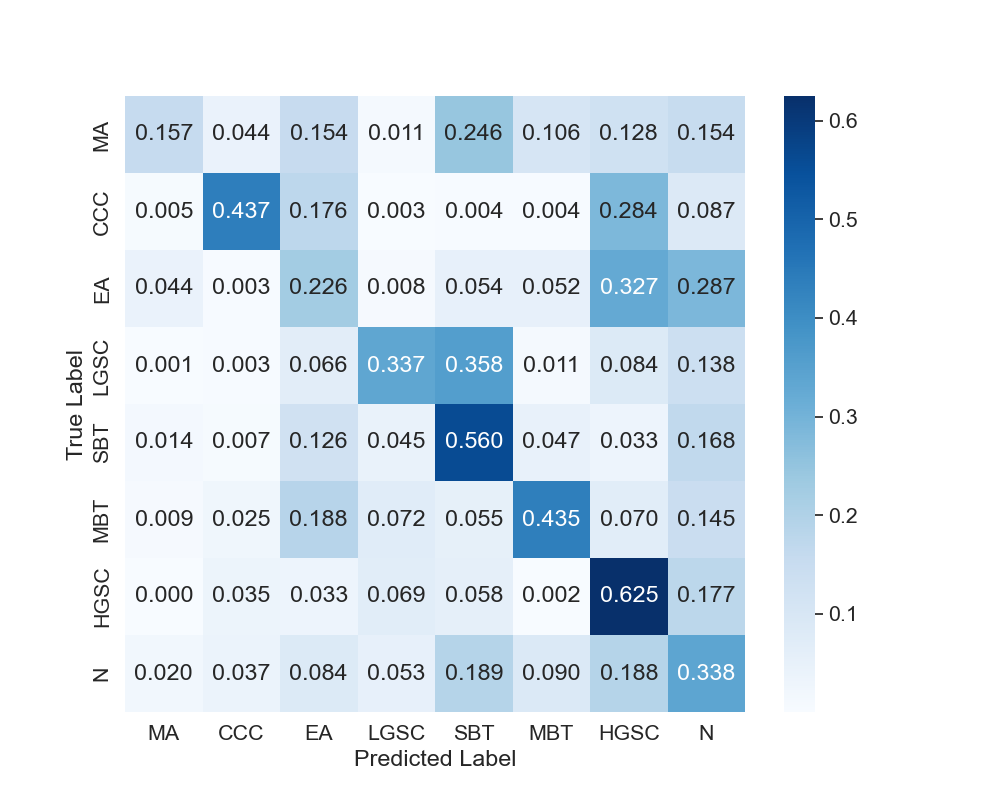}
\caption{}
\end{subfigure}
\begin{subfigure}[H]{0.49\textwidth}
\centering
\includegraphics[width=\textwidth]{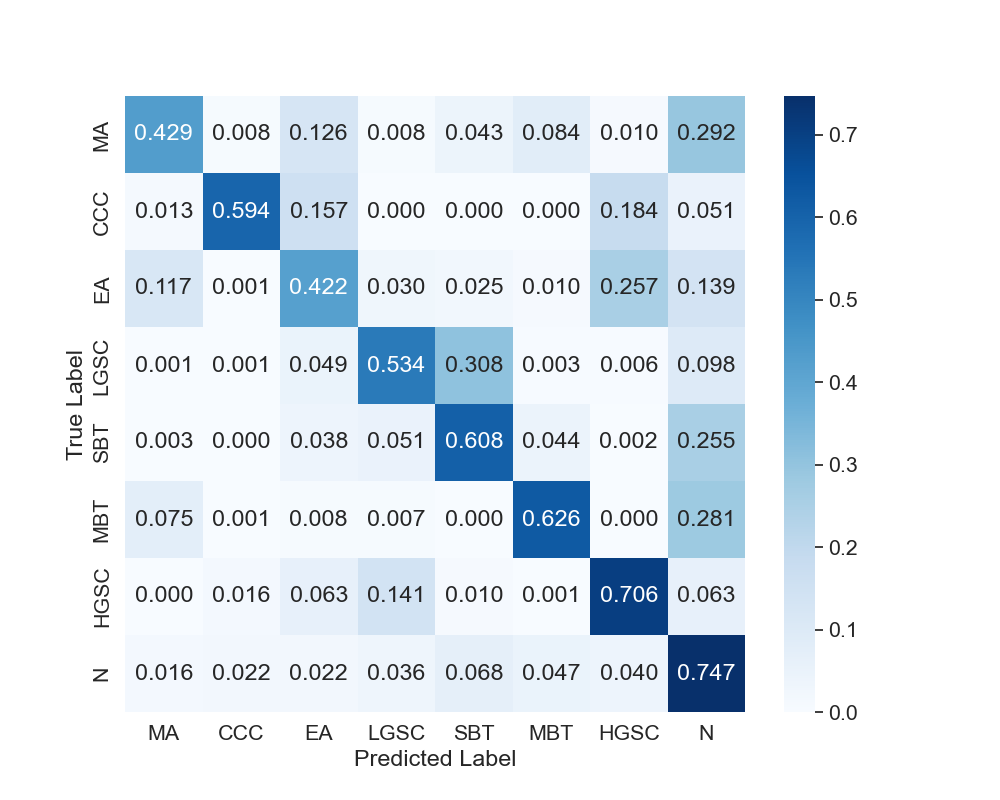}
\caption{}
\end{subfigure}

\begin{subfigure}[H]{0.49\textwidth}
\centering
\includegraphics[width=\textwidth]{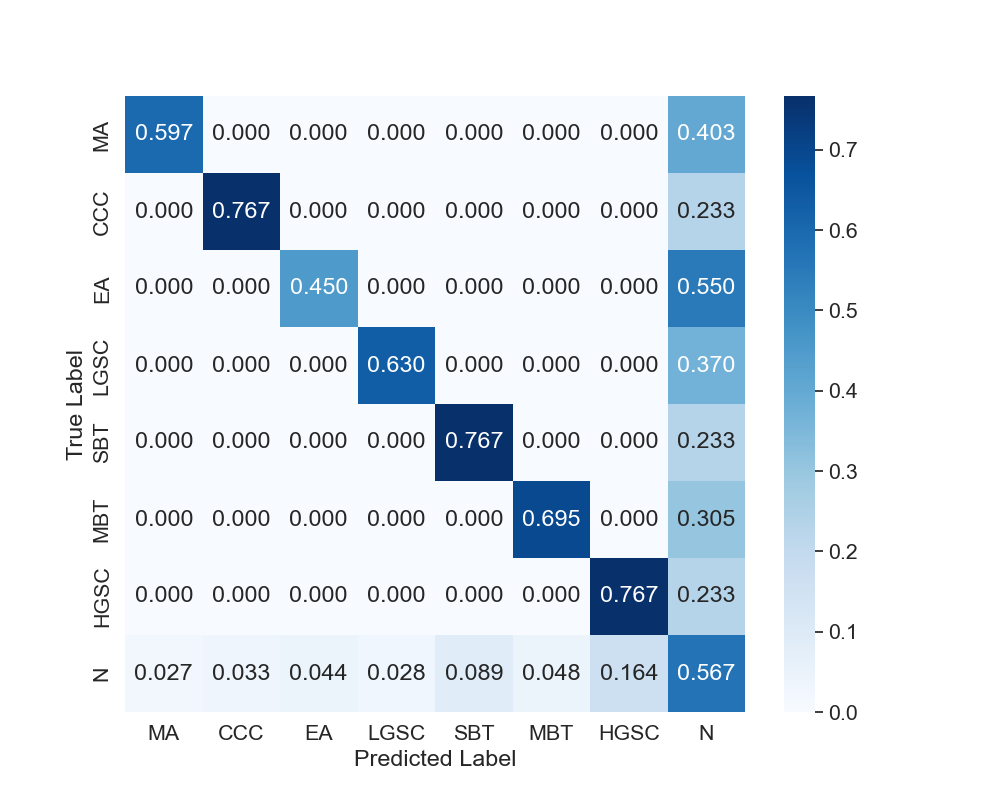}
\caption{}
\end{subfigure}
\begin{subfigure}[H]{0.49\textwidth}
\centering
\includegraphics[width=\textwidth]{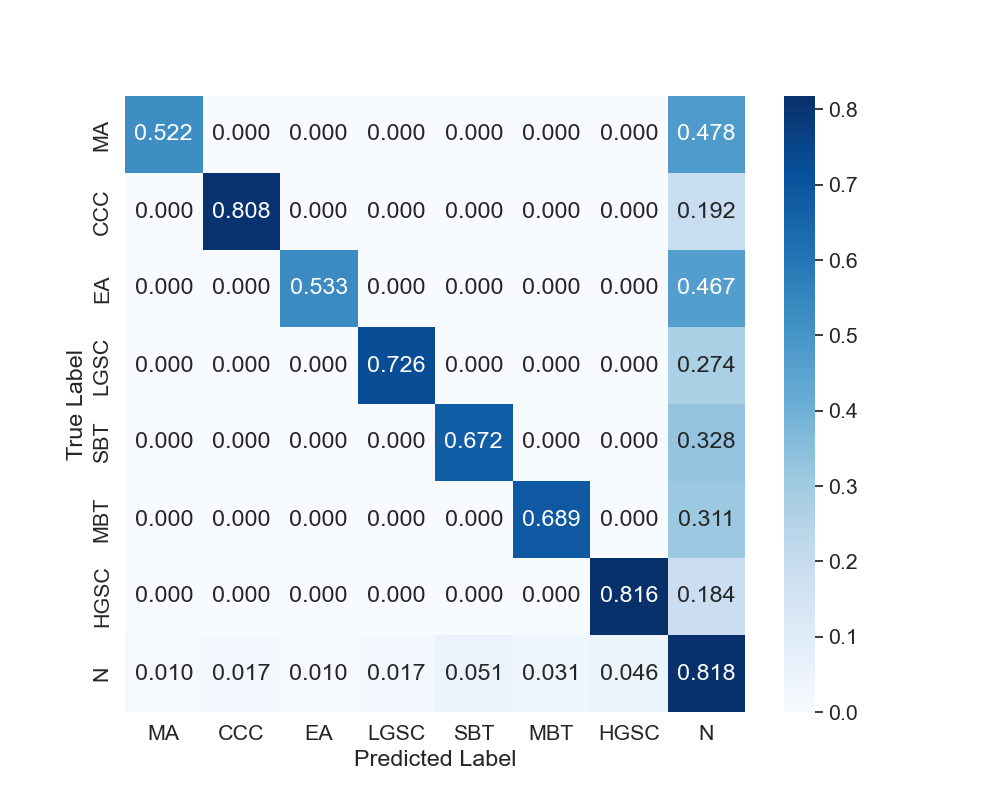}
\caption{}
\end{subfigure}

\caption{Normalised confusion matrices for slide classifications for (a) DSMIL and (b) MB-DSMIL-CL-PL, for instance classifications for (c) DSMIL and (d) MB-DSMIL-CL-PL, and for filtered instance classifications for (e) DSMIL and (f) MB-DSMIL-CL-PL.} \label{fig:confusion_matrices}
\end{figure}


\bibliographystyle{agsmdoi}
\bibliography{references.bib}






\end{document}